\documentclass[11pt,letterpaper]{article}

\usepackage[T1]{fontenc}
\usepackage{fullpage}
\usepackage{lmodern}
\usepackage[round,authoryear]{natbib}
\usepackage{iftex}
\ifPDFTeX
  \usepackage[protrusion=true,expansion=true]{microtype}
\else
  \usepackage[protrusion=true,expansion=false]{microtype}
\fi
\usepackage{amsmath,amsfonts,bm}

\def\eqref#1{equation~\ref{#1}}

\def\1{\bm{1}}

\DeclareMathAlphabet{\mathsfit}{\encodingdefault}{\sfdefault}{m}{sl}
\SetMathAlphabet{\mathsfit}{bold}{\encodingdefault}{\sfdefault}{bx}{n}

\usepackage{amsmath}
\usepackage{amssymb}
\usepackage{booktabs}
\usepackage{caption}
\usepackage{array,tabularx}
\usepackage{placeins}
\usepackage{needspace}
\usepackage{graphicx}
\usepackage{pgfplots}
\pgfplotsset{compat=1.18}
\usepgfplotslibrary{groupplots}
\usetikzlibrary{arrows.meta,positioning,calc,fit}
\usepackage{xcolor}
\usepackage{inconsolata}
\usepackage{fontawesome5}
\usepackage{xurl}
\usepackage[colorlinks,linkcolor=red,anchorcolor=blue,citecolor=blue]{hyperref}
\usepackage{titletoc}

\numberwithin{equation}{section}
\renewcommand{\arraystretch}{1.15}
\titlecontents{appsection}
  [15pt]
  {\addvspace{8pt}\normalfont\bfseries}
  {\contentslabel[\thecontentslabel\hfill]{15pt}}
  {\hspace*{-15pt}}
  {\hfill\contentspage}
\titlecontents{appsubsection}
  [39pt]
  {\addvspace{2pt}\normalfont}
  {\contentslabel[\thecontentslabel\hfill]{24pt}}
  {\hspace*{-24pt}}
  {\titlerule*[8pt]{.}\contentspage}

\newcommand{\GRPO}{GRPO}
\newcommand{\PRM}{PRM}
\newcommand{\OPSD}{OPSD}
\newcommand{\GRPOPRM}{\GRPO{}+\PRM{}}
\newcommand{\outcomeGRPO}[1][ ]{outcome-only#1\GRPO{}}
\newcommand{\OutcomeGRPO}[1][ ]{Outcome-only#1\GRPO{}}
\newcommand{\GRPOPRMFixed}[1][ ]{\GRPOPRM{}#1(fixed~\ensuremath{\lambda})}
\newcommand{\GRPOPRMDecayed}[1][ ]{\GRPOPRM{}#1(decayed~\ensuremath{\lambda})}
\newcommand{\OPSDReference}[1][ ]{\OPSD{}#1(\referenceAnswers{})}
\newcommand{\OPSDSuggestions}[1][ ]{\OPSD{}#1(\turnSuggestions{})}

\newcommand{\privilegedInformation}{privileged information}
\newcommand{\PrivilegedInformation}{Privileged information}
\newcommand{\referenceAnswers}{reference answers}
\newcommand{\ReferenceAnswers}{Reference answers}
\newcommand{\turnSuggestions}{turn-level suggestions}
\newcommand{\TurnSuggestions}{Turn-level suggestions}

\newcommand{\processRewardWeight}{process-reward weight}
\newcommand{\ProcessRewardWeight}[1][ ]{Process-reward#1weight}

\newcommand{\baseModel}{Qwen3.5-9B}

\newcommand{\lei}[1]{}

\title{When Fine-Grained Credit Assignment Helps and Hurts: \\
Multimodal Agentic Reinforcement Learning through Video Editing}

\author{Lei Zhao$^{1}$\thanks{%
Part of this work was completed while Lei Zhao was an intern at Philo Labs.
Emails, in author order: \texttt{leizhao7@upenn.edu},
\texttt{tom@philolabs.ai}, and \texttt{christine@philolabs.ai}.}%
\quad Zongheng Cao$^{2}$ \quad Xinyu Hu$^{2}$\\[0.6em]
{\normalsize $^{1}$University of Pennsylvania \quad $^{2}$Philo Labs}}
\hypersetup{pdfauthor={Lei Zhao, Zongheng Cao, Xinyu Hu}}

\date{}

\begin{document}

\maketitle

\begin{abstract}
Reinforcement learning (RL) has shown promise for training multimodal agents to reason and use tools. However, it remains unclear when fine-grained credit assignment improves learning and when it fails to improve task performance. Video editing provides a useful testbed for this question, combining multimodal perception, long-horizon planning, and tool use with delayed outcome feedback. We construct four movie-trailer editing task types---Content Selection, Narrative Sequencing, Rough Cut, and Fine Cut---that reflect real-world editing workflows and span different levels of task complexity and interaction horizon. Decaying process rewards provides the largest benefit on Fine Cut, with a 45.4\% relative gain over \outcomeGRPO{} on tasks averaging more than 60 turns under \outcomeGRPO{}. On the three shorter-horizon task types averaging 30--40 turns under the same baseline, \outcomeGRPO{} remains competitive, with decaying process rewards yielding at most a 10.4\% relative gain. Process rewards from an LLM judge initially improve tool use and encourage visual inspection. As training progresses, however, GRPO+PRM with a fixed \processRewardWeight{} increasingly exhibits excessive reasoning and redundant tool calls: process rewards continue to rise while task completion rates and outcome scores decline. Decaying the \processRewardWeight{} to zero mitigates this reward-hacking behavior while preserving the early gains. Despite dense token-level supervision, on-policy self-distillation (\OPSD{}) yields little improvement or substantial degradation and is highly sensitive to the choice of \privilegedInformation{}. Our analysis reveals differences in teacher--student disagreement and gradient magnitudes, and suggests possible dilution of informative signals over long trajectories.

\par\medskip
\begingroup
\definecolor{resourceurl}{HTML}{C24685}
\hypersetup{urlcolor=resourceurl}
\urlstyle{tt}
\setlength{\tabcolsep}{2pt}
\renewcommand{\arraystretch}{1.12}
\noindent
\begin{tabularx}{\linewidth}{@{}p{1.4em}>{\bfseries\sffamily}p{7.1em}>{\raggedright\arraybackslash}X@{}}
\textcolor{resourceurl}{\faGlobe} & Project Page: & \url{https://leizhao7.github.io/videoedit/} \\
\faGithub & GitHub: & \url{https://github.com/leizhao7/video-editing-credit-assignment} \\
\raisebox{-0.22em}{\includegraphics[width=1.15em]{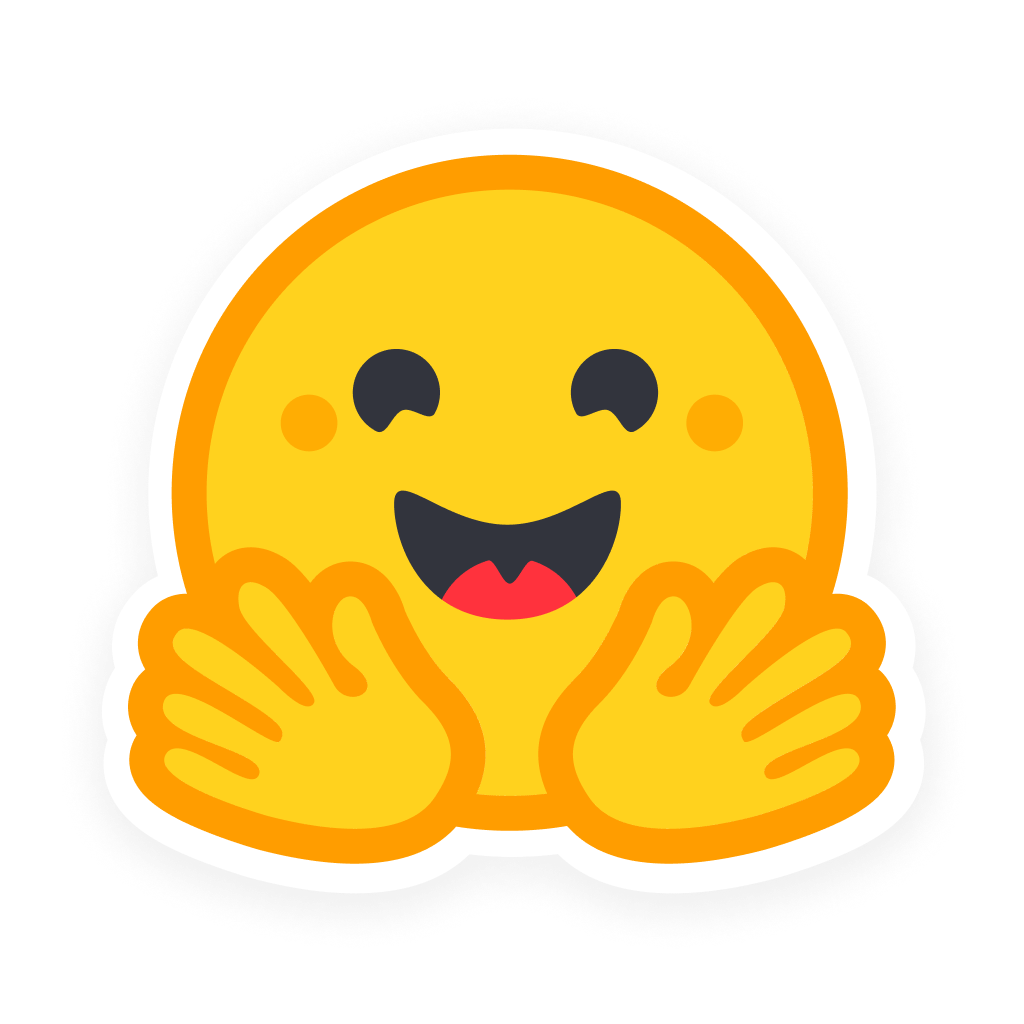}} & Models: & \href{https://huggingface.co/collections/leizhao7/when-fine-grained-credit-assignment-helps-and-hurts}{\nolinkurl{https://huggingface.co/collections/leizhao7/}}
\newline\href{https://huggingface.co/collections/leizhao7/when-fine-grained-credit-assignment-helps-and-hurts}{\nolinkurl{when-fine-grained-credit-assignment-helps-and-hurts}} \\
\end{tabularx}
\endgroup
\end{abstract}

\begin{figure}[!t]
  \centering
  \includegraphics[width=\linewidth]{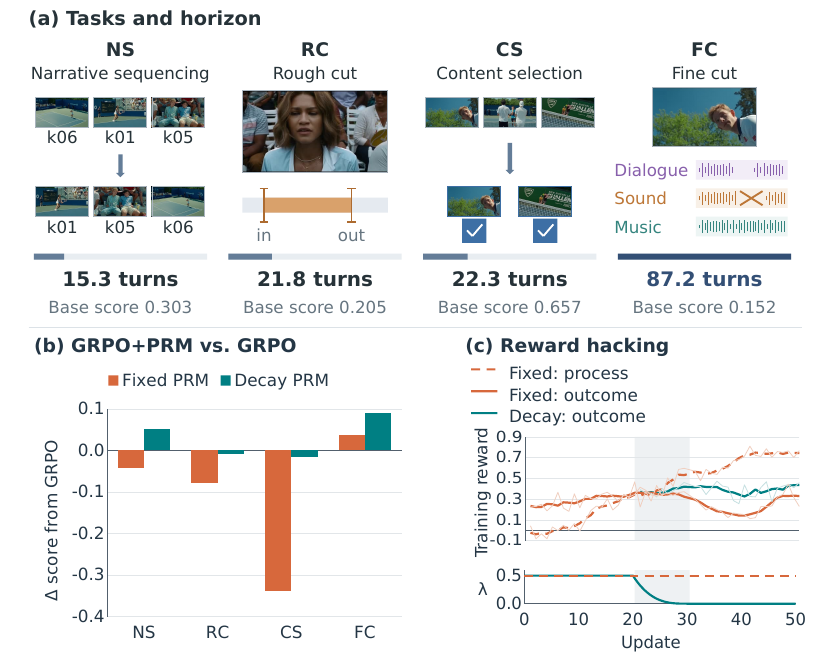}
  \caption{\textbf{Task-dependent effects of process rewards for \baseModel{}.}
  (a) Independent task schematics, ordered by base-policy horizon. Turns and base
  scores are task-type averages, not example-specific results; evaluation metrics
  average four repetitions. Fine Cut keeps picture and timing fixed.
  (b) Final score differences from \outcomeGRPO{}.
  (c) Fixed-weight process and outcome rewards, and decayed-weight outcome rewards
  available from update 21. Faint lines connect recorded observations; thick lines
  show centered five-observation means. The aligned strip shows fixed and
  cubic-decay weights $\lambda$. Shading marks the decayed variant's 20--30 decay window.}
  \label{fig:teaser}
\end{figure}

\section{Introduction}
\label{sec:introduction}

Video agents are moving from video understanding towards long-horizon multimodal tasks
such as video production, where models must integrate multimodal perception, planning,
and tool
use~\citep{wang2024lave,yang2024avt,sandoval2025editduet,zhao2026cutclaw,zhang2026cineagents}.
However, even the strongest agent systems struggle on these
tasks~\citep{cao2026agenticvbench,hu2026cutverse}. Reinforcement learning has shown
promise for improving their performance, enabling agents to refine their decision-making
and tool-use policies through feedback from task
execution~\citep{kulkarni2025avatar,zhang2025vital,yang2025longvt,zhang2026eva,wang2025ragen}.
However, reward formulation remains challenging because the quality of the resulting
video rests on subjective judgment~\citep{yan2026crayotter,wen2026avecompass}. In
addition, feedback on the final video becomes available only after a sequence of
interdependent editing decisions, making it difficult to determine how individual
actions contribute to the final outcome. Credit assignment is therefore central to
training these agents~\citep{arjonamedina2018rudder,feng2025gigpo,tan2026hcapo}. Yet it
remains unclear when finer-grained credit assignment helps or hurts learning in
long-horizon multimodal tasks.

Most work on video production builds a workflow around a model that it never
updates~\citep{sandoval2025editduet,zhao2026cutclaw,zhang2026cineagents}. Crayotter trains an editing agent with segment-level credit,
where segments group consecutive editing events from the same stage~\citep{yan2026crayotter}. We ask how
trajectory-, turn-, and token-level credit compare across editing tasks under a
shared training setup. Elsewhere, several lines report that finer credit helps. VAGEN gives a multi-turn visual agent dense turn-level
supervision~\citep{wang2025vagen}, TAPO repairs the credit that a trajectory-level
advantage misassigns to tool calls inside failed rollouts~\citep{dong2026tapomm}, and
interface agents trained on step-level progress signals beat actors trained on outcome
rewards alone~\citep{zhang2025progrm,chen2025guishepherd}. Other work reports negative
results for process rewards. Group-relative optimization with an outcome reward provably
carries an implicit process reward, which makes an explicit one look
redundant~\citep{sullivan2025grpoprm}. A process reward is dense but remains prone to
bias and reward hacking~\citep{zheng2026admire,pan2026rmsurvey}. Outcome-only training
is argued to suffice on long horizons once exploration is wide
enough~\citep{pu2026driftless}. How credit should be assigned in long-horizon multimodal
agents therefore remains unclear.

We study the question in movie-trailer editing. We generate 797 tasks from 49 real
movies and their released trailers. The work decomposes into four stages that a human
editor performs in order, which give the task types Content Selection (CS), Narrative
Sequencing (NS), Rough Cut (RC), and Fine Cut (FC). The four types differ in editing requirements, interaction length, and in
the quality profile of the turns they elicit, which lets us
ask how the same credit signal behaves as the task changes. We train
\baseModel{}~\citep{qwen2026qwen35} for 50 updates on 716 of these tasks under four
schemes that cover the three granularities. At the trajectory level we use \outcomeGRPO{}. At the turn level we compare \GRPOPRMFixed{} and
\GRPOPRMDecayed{}. Both use a process reward from Gemini 3.8 Flash,
averaged over three votes. At the token level we use on-policy self-distillation (\OPSD{}).
Its teacher receives \privilegedInformation{} that the student never sees:
\referenceAnswers{} or \turnSuggestions{}. Outcome scores come from Gemini 3.1 Pro, so the process reward
and the outcome reward never share a model. The schemes share the base model, training tasks, and number of updates,
while objective-specific settings are detailed in Appendix~\ref{app:optimization}. We evaluate every run
on 81 validation tasks four times and report the mean, and we record how the agent
behaves as well as what it scores.

Our contributions are threefold.
First, we develop a data-construction pipeline that turns films and their
released trailers into four types of editing tasks reflecting real-world
editing workflows. Training on these tasks improves \baseModel{}'s
video editing performance on held-out films.
Second, we compare trajectory-, turn-, and token-level learning signals under
a shared setup. Decaying process rewards yields a 45.4\% relative gain over
\outcomeGRPO{} on Fine Cut, while \outcomeGRPO{} remains competitive on the
three shorter-horizon task types.
Third, we analyze training dynamics and model behavior to identify limitations
of the tested finer-grained signals: fixed process rewards can undermine task
completion despite encouraging desirable behaviors, and \OPSD{} is sensitive
to the \privilegedInformation{} provided to its teacher.

\section{Related Work}
\label{sec:related-work}

\paragraph{Agents for video production.}
Video-editing agents have evolved from executing language-guided actions to managing long-horizon multi-stage production~\citep{wang2024lave,yang2024avt,sandoval2025editduet,wang2025hive,ding2025promptedit,zhao2026cutclaw,li2026direct,zhang2026cineagents,liang2025univa,yan2026crayotterworkflow,zhou2026videoagent}.
These systems focus on designing editing workflows while keeping the underlying models fixed.
Despite this progress, advanced agents remain unreliable in professional video post-production tasks~\citep{cao2026agenticvbench,hu2026cutverse}.
Another line of work trains video-editing policies~\citep{tang2022msan,hu2024rlve,zhou2026autocut,yu2026aurora}.
These methods use policy gradients for shot composition or supervised and preference-based post-training for tool-using planners.
Crayotter assigns segment-level credit based on rankings of finished videos produced for the same task~\citep{yan2026crayotter}.
Our study compares trajectory-, turn-, and token-level supervision across a shared suite of editing tasks.

\paragraph{Credit assignment for long-horizon agents.}
Attributing delayed outcomes to the decisions that caused them is a long-standing
challenge~\citep{arjonamedina2018rudder,ren2021rrd,harutyunyan2019hca,mesnard2020cca,
meulemans2023cocoa}.
Research on language agents builds explicit learning signals at several granularities.
Trajectory-level methods score a whole rollout against its
peers~\citep{shao2024deepseekmath,wang2025ragen}.
Turn-level methods estimate value, progress, or a relative advantage for each
turn~\citep{zhou2024archer,wang2025sparl,feng2025gigpo,xi2025agentprm,li2025turnppo,
tan2026hcapo,liu2025istar}.
Finer methods assign an advantage to a reasoning step or a segment within a
response~\citep{kazemnejad2024vineppo,guo2025spo}.
Multimodal work follows the same three levels across visual, interface, embodied, and video
agents~\citep{wang2025vagen,pang2026ica,dong2026tapomm,tan2025tarl,zheng2026admire,
chen2025guishepherd,zhang2025progrm,zhou2026tgrpo,xu2025starevla,kulkarni2025avatar,
li2026timethink,zhang2025vital,yang2025longvt,zhang2026eva}.
Whether explicit finer-grained supervision improves learning remains an empirical question.
Under mild assumptions, \outcomeGRPO{} can be written as a process-aware objective that
carries an implicit Monte Carlo process reward~\citep{sullivan2025grpoprm}.
Process reward models are also hard to build and to evaluate. Synthesizing their labels by
Monte Carlo estimation generalizes worse than using a judge or a human
annotator~\citep{zhang2025prmlessons}, and benchmarks for fine-grained error detection
expose substantial weaknesses in current models~\citep{song2025prmbench,sun2026grben}.
Empirical studies nevertheless report benefits from denser
feedback~\citep{wang2025practitioners,wu2026demystifying}, alongside failure modes
involving teacher guidance, reward-model overoptimization, and proxy
exploitation~\citep{li2026rethinkingopd,fu2026revisitingopd,skalse2022rewardhacking,
gao2022overopt,thaman2026rhbench,pu2026driftless,gunjal2025rubrics,shao2025drtulu}.
Evidence remains limited on when finer-grained supervision helps or hurts long-horizon
multimodal agents, and how these effects vary across tasks and training stages.

\section{Preliminaries}
\label{sec:preliminaries}

\subsection{Long-Horizon Multimodal Agent}

We consider a task $x$ drawn from a task distribution $\mathcal{D}$, in which an
agent interacts with an environment over many turns, following standard formulations of long-horizon
agentic RL~\citep{zhou2024archer,wang2025ragen,feng2025gigpo,zhang2026eva}. At turn
$t$ the agent observes a state $s_t\in\mathcal{S}$ and emits an action
$a_t\in\mathcal{V}^*$, and the environment returns the next state,
\begin{equation}
    a_t\sim\pi_\theta(\cdot\mid s_t,x),
    \qquad
    s_{t+1}\sim\mathcal{P}(\cdot\mid s_t,a_t).
\end{equation}
The state $s_t$ carries images, tool outputs, and a mutable workspace alongside the
textual history. The episode yields a trajectory
$\tau=(s_0,a_0,\ldots,a_{T-1},s_T)$ and a delivered artifact
$y_\tau=\operatorname{Output}(s_T)$, which a grader maps to a terminal reward
$R_{\mathrm{out}}(y_\tau)\in[0,1]$. No reward is available before the agent commits.

\subsection{Credit Assignment}

We next present three credit assignment schemes at different granularities.

\paragraph{Trajectory level.}
At this granularity we study Group Relative Policy Optimization
(\GRPO{}) \citep{shao2024deepseekmath} with an outcome-only reward. \GRPO{} estimates
advantages from a group of rollouts without a learned critic. For each task $x$, we
sample $G$ trajectories $\{\tau_i\}_{i=1}^G$ from $\pi_{\mathrm{old}}$ and score them
with $o_i=R_{\mathrm{out}}(y_{\tau_i})$. The group-normalized advantage is
\begin{equation}
    A_i^{\mathrm{out}}=\frac{o_i-\mu_o}{\sigma_o+\epsilon},
    \label{eq:grpo-advantage}
\end{equation}
with $\mu_o$ and $\sigma_o$ the group mean and sample standard deviation.
We optimize the policy using rollout importance sampling with token-mean reduction
over valid assistant tokens (Appendix~\ref{app:experimental-details}). Every token of
$\tau_i$ receives the same $A_i^{\mathrm{out}}$, so the signal does not vary within a
trajectory. We write this setting as \outcomeGRPO{}.

\paragraph{Turn level.}
A process reward model (\PRM{}) scores each action from its interaction history and the
tool feedback it produced~\citep{wang2026openclawrl}. We average the $M$ valid judgments of a turn into $r_{i,t}$
and add a normalized process term to the shared outcome term,
\begin{equation}
    A_{i,t}=A_i^{\mathrm{out}}+\lambda A_{i,t}^{\mathrm{proc}},
    \qquad
    A_{i,t}^{\mathrm{proc}}=\frac{r_{i,t}-\mu_t}{\sigma_t+\epsilon},
    \label{eq:prm-advantage}
\end{equation}
where $\mu_t$ and $\sigma_t$ are the mean and sample standard deviation of valid
process scores at the same turn index $t$ across rollouts of task $x$.
We do not apply additional batch-level advantage normalization.
We call $\lambda$ the \processRewardWeight{} and compare two settings.
\GRPOPRMFixed{} keeps this weight fixed throughout training.
\GRPOPRMDecayed{} reduces it to zero partway through training.
Later updates therefore use only the outcome reward. Every token of action $a_{i,t}$ receives
$A_{i,t}$, so the signal varies across turns but not within one.

\paragraph{Token level.}
On-policy self-distillation (\OPSD{}) matches a teacher distribution on the student's own
prefixes~\citep{zhao2026self}. The teacher $\pi_{\mathrm{tea}}$ is a frozen copy
of the initial policy and receives \privilegedInformation{} $c_t$ that the student never
sees. We compare two forms of \privilegedInformation{}: \referenceAnswers{} and
\turnSuggestions{}. For token $k$ of action $a_t$ we write
$p_{t,k}(v)=\pi_\theta(v\mid s_t,x,a_{t,<k})$ for the student and
$q_{t,k}(v)=\pi_{\mathrm{tea}}(v\mid s_t,x,c_t,a_{t,<k})$ for the teacher. All \OPSD{} experiments use the forward KL divergence,
\begin{equation}
    \mathcal L(\theta)
       =\mathbb E_{t,k}\!\left[D_{\mathrm{KL}}(q_{t,k}\|p_{t,k})\right],
    \label{eq:opsd-kl}
\end{equation}
taken over the generated assistant tokens. In practice we evaluate the KL on the
teacher's top-$K$ support rather than the full vocabulary, and
Appendix~\ref{app:experimental-details} gives the details. \OPSD{} trains on this loss with no
reward term at all. Every token receives its own target, so the signal varies within a
turn as well.

\section{Task Suite}
\label{sec:movie-trailer-editing}

\subsection{Tasks and Environment}
\label{sec:trailer-problem-setup}

We study movie-trailer editing as it is done in
practice~\citep{murch2001blink,pearlman2009cutting}. We decompose the editing of one
trailer into four stages that a human editor works through in order, following the
workflow-oriented view of AgenticVBench~\citep{cao2026agenticvbench}. Each stage gives
one task type, namely Content Selection (CS), Narrative Sequencing (NS), Rough Cut (RC), and Fine Cut (FC).
Figure~\ref{fig:teaser}(a) illustrates the four editing operations.
A task consists of a multi-turn episode in which the agent inspects the supplied media
and commits its artifact. The released trailer provides a reference answer for each
task, held by the evaluator and hidden from the agent.
Table~\ref{tab:trailer-task-stages} gives the task, the input, and the output of each
type of task. See Appendix~\ref{app:task-construction} for the definition of each type and
Appendix~\ref{app:example-tasks} for two worked examples. Base scores range from 0.152 to 0.657 under the respective task metrics,
and mean interaction horizons range from 15.3 to 87.2 turns.

\paragraph{Scoring.}
Each task is scored on the artifact the agent commits when the episode ends. All tasks use the
same three channels for this outcome score $R_{\mathrm{out}}$,
\begin{equation}
    R_{\mathrm{out}} = w_{\mathrm{det}}R_{\mathrm{det}}
        + w_{\mathrm{ref}}R_{\mathrm{ref}}
        + w_{\mathrm{blind}}R_{\mathrm{blind}},
    \label{eq:outcome-reward}
\end{equation}
where $R_{\mathrm{det}}$ is a deterministic verifier that scores the artifact against
that reference answer, $R_{\mathrm{ref}}$ is a multimodal LLM judge that scores the
artifact with that reference, and $R_{\mathrm{blind}}$ is the same judge asking whether
the artifact works on its own terms. Both judge channels score against a weighted
rubric, and one model call grades each rubric item with its own evidence attached. The weights of Equation~\ref{eq:outcome-reward} sum to one and vary with the
type. In this paper, we use Gemini 3.1 Pro as the multimodal LLM judge. The process rewards
used during training come from a different model.

Content Selection and Narrative Sequencing use only the deterministic score, with weights $(1,0,0)$.
Content Selection scores the unordered selected shot set $S$ against the reference
set $S^\star$ using F1, $2|S\cap S^\star|/(|S|+|S^\star|)$.
It evaluates selection alone, with partial credit for overlap.\footnote{Selecting every candidate would yield a mean F1 of 0.478 on the 14 Content Selection validation tasks.}
Narrative Sequencing scores
agreement with the reference order. Rough Cut uses $(0.4,0.4,0.2)$, combining
deterministic cut-point similarity with reference-conditioned and reference-free
judge scores. Fine Cut uses $(0,0.4,0.6)$ to assess sound design and transitions
through the two judge channels. These task-specific metrics measure different
editing objectives, so their absolute values are not a common scale of task difficulty.
Table~\ref{tab:rubrics} lists the judge rubrics, and
Appendix~\ref{app:experimental-details} gives the aggregation.

\paragraph{Editing environments and harness.}
Each task runs in a fresh workspace holding the editorial brief, a manifest, and the
media the task type exposes. The agent is provided with a shell and eight media tools for
inspecting pictures, probing audio, and validating its own artifact. See
Appendix~\ref{app:tools} for the full list. We run the agent in the official OpenCode
harness. To keep training and rollout on the same pixels, we add a plugin to the harness
that registers every image before it enters the conversation. See
Appendix~\ref{app:editing-environment} for details.

\begin{table}[htbp]
\caption{Task stages, inputs and outputs of the four task types. CS: Content Selection; NS: Narrative Sequencing; RC: Rough Cut; FC: Fine Cut. Each task also receives an editorial
brief. The RC target specifies total duration. Turns and score are averaged over
four independent evaluations of all 81 validation tasks with the untrained \baseModel{}.}
\label{tab:trailer-task-stages}
\setlength{\tabcolsep}{3pt}
\begin{center}
\begin{tabularx}{\linewidth}{@{}>{\raggedright\arraybackslash}p{76pt}>{\raggedright\arraybackslash}p{44pt}>{\raggedright\arraybackslash}X>{\raggedright\arraybackslash}p{72pt}rr@{}}
\toprule
Task stage & Task type & Input & Output & Turns & Score \\
\midrule
Understanding & CS & Candidate shots & Selected shot IDs & 22.3 & 0.657 \\
\addlinespace[2pt]
 & NS & Shuffled shots & Ordered shot IDs & 15.3 & 0.303 \\
\midrule
Editing & RC & Ordered shots, target duration & Clip in/out points & 21.8 & 0.205 \\
\addlinespace[2pt]
 & FC & Rough cut, audio assets & Rendered edit & 87.2 & 0.152 \\
\bottomrule
\end{tabularx}
\end{center}
\end{table}

\subsection{Data Construction and Dataset}
\label{sec:trailer-task-construction}
\label{sec:trailer-data-filtering}

Given a source film and its released trailer, we detect shot boundaries and align
usable trailer shots to their source-film footage. We partition the trailer into
narratively coherent chunks and generate an editorial brief for each, describing
its narrative role without identifying the clips. Each chunk yields one task of
each type by varying the supplied media and required editing operation
(Table~\ref{tab:trailer-task-stages}). Figure~\ref{fig:pipeline} illustrates the
pipeline, and Appendix~\ref{app:task-construction} gives the construction details.

The pipeline produces 2,352 tasks from 49 films. After filtering with Gemini 3.8 Flash we
sample 797 of them, 716 for training from 43 films and 81 for validation from 6 films that
do not appear in training. Table~\ref{tab:filtered-dataset} gives the split by task type,
and Appendix~\ref{app:task-construction} lists the films and describes the audit and the
sampling protocol.

\begin{table}[htbp]
\caption{Task counts by type. The training and validation films are disjoint.}
\label{tab:filtered-dataset}
\begin{center}
\begin{tabular*}{\linewidth}{@{}l@{\extracolsep{\fill}}rrrrr@{}}
\toprule
Split & CS & NS & RC & FC & Total \\
\midrule
Train & 122 & 132 & 231 & 231 & 716 \\
Validation & 14 & 15 & 26 & 26 & 81 \\
\bottomrule
\end{tabular*}
\end{center}
\end{table}

\section{Experiments}
\label{sec:experiments}

\subsection{Experimental Setup}
\label{sec:experimental-setup}

\paragraph{Training settings.}
We train \baseModel{}~\citep{qwen2026qwen35} for 50 updates on the 716 training
tasks (Section~\ref{sec:trailer-data-filtering}) under four different schemes
for credit assignment: (i) \outcomeGRPO{}, (ii) \GRPOPRMFixed{},
(iii) \GRPOPRMDecayed{}, and (iv) \OPSD{}.
For \GRPOPRMFixed{}, we use a \processRewardWeight{} of $\lambda=0.5$ throughout training.
\GRPOPRMDecayed{} keeps the same weight until update 20.
We then decay it to zero over updates 20--30 and keep it at zero thereafter.
Section~\ref{sec:analysis-decay} explains the choice of this interval, and
Appendix~\ref{app:experimental-details} gives the decay schedule and full training configuration.
For \OPSD{}, we compare two forms of \privilegedInformation{}: \referenceAnswers{} and
\turnSuggestions{}. Gemini 3.8 Flash generates the \turnSuggestions{} from the
student's context at each turn. These suggestions are provided to the frozen teacher as
\privilegedInformation{}.

\paragraph{Evaluation.}
We evaluate the final checkpoint of each run on the 81 validation tasks
introduced in Section~\ref{sec:trailer-data-filtering}. To reduce evaluation
variance, we evaluate every task four times and report the mean score, denoted
avg@4. See Appendix~\ref{app:experimental-details} for the full evaluation
configuration.

\subsection{Experiment Results}
\label{sec:main-results}

Table~\ref{tab:main-results} summarizes the results.
\OutcomeGRPO{} improves all four task types, raising the overall score from 0.284 to 0.413.
Unexpectedly, \GRPOPRMFixed{} scores lower than \outcomeGRPO{} on every task type except Fine Cut.
Decaying the \processRewardWeight{} improves all four task scores over \GRPOPRMFixed{}
and yields the highest overall score of 0.447.
Its largest gain over \outcomeGRPO{} is on Fine Cut (0.293 vs.\ 0.201).
\OPSDReference{} performs similarly to the untrained model (0.277 vs.\ 0.284 overall),
while \OPSDSuggestions{} drops to 0.044.
Section~\ref{sec:analysis} analyzes these contrasting outcomes.

\begin{table}[htb]
\caption{Validation scores for the final checkpoints of all training variants across the four task types. Each score is the mean over four repetitions (avg@4). Percentages below scores show relative changes from the base model, computed using the four-decimal evaluation means. Best result per column in bold.}
\label{tab:main-results}
\setlength{\tabcolsep}{4pt}
\begin{center}
\begin{tabular}{@{}lccccc@{}}
\toprule
& \multicolumn{5}{c}{Mean task score (avg@4)} \\
\cmidrule(lr){2-6}
Method & CS & NS & RC & FC & Overall \\
\midrule
\baseModel{} & 0.657 & 0.303 & 0.205 & 0.152 & 0.284 \\
\midrule
\OutcomeGRPO{} & \textbf{0.821} & 0.500 & \textbf{0.356} & 0.201 & 0.413 \\
& {$(+24.9\%)$} & {$(+65.1\%)$} & {$(+73.4\%)$} & {$(+32.9\%)$} & {$(+45.5\%)$} \\
\midrule
\GRPOPRMFixed{} & 0.482 & 0.457 & 0.276 & 0.239 & 0.334 \\
& {$(-26.6\%)$} & {$(+51.0\%)$} & {$(+34.8\%)$} & {$(+57.8\%)$} & {$(+17.4\%)$} \\
\GRPOPRMDecayed{} & 0.804 & \textbf{0.552} & 0.348 & \textbf{0.293} & \textbf{0.447} \\
& {$(+22.3\%)$} & {$(+82.2\%)$} & {$(+69.5\%)$} & {$(+93.2\%)$} & {$(+57.2\%)$} \\
\midrule
\OPSDReference{} & 0.606 & 0.267 & 0.219 & 0.162 & 0.277 \\
& {$(-7.8\%)$} & {$(-11.7\%)$} & {$(+6.8\%)$} & {$(+6.9\%)$} & {$(-2.7\%)$} \\
\OPSDSuggestions{} & 0.129 & 0.063 & 0.032 & 0.000 & 0.044 \\
& {$(-80.4\%)$} & {$(-79.2\%)$} & {$(-84.5\%)$} & {$(-100.0\%)$} & {$(-84.4\%)$} \\
\bottomrule
\end{tabular}
\end{center}
\end{table}

\section{Credit Assignment: Benefits and Failure Modes}
\label{sec:analysis}

We examine trajectory-level credit on long horizons in Section~\ref{sec:analysis-credit}.
Section~\ref{sec:analysis-hacking} links process-reward hacking to declining task completion and studies \processRewardWeight{} decay as a mitigation.
Section~\ref{sec:analysis-opsd} examines the sensitivity of \OPSD{} to \privilegedInformation{}.

\subsection{Task Structure and Trajectory-Level Credit}
\label{sec:analysis-credit}
\paragraph{Gains vary across editing tasks.}
As shown in Table~\ref{tab:main-results}, \outcomeGRPO{} raises the overall
score from 0.284 to 0.413. However, Fine Cut improves by only 0.050,
whereas the other three task types each improve by at least 0.15.
On Fine Cut, adding process rewards further raises the score from
0.201 to 0.239 with a fixed weight and 0.293 with a decaying weight.
These improvements show that further gains on Fine Cut are achievable
beyond the outcome-only result.
On the other three task types, the tested finer-grained methods yield
limited additional gains over \outcomeGRPO{}: none improves Content
Selection or Rough Cut, and the largest gain on Narrative Sequencing
is 0.052.

\paragraph{A possible explanation: long horizons mix good and bad turns.}

To understand why \outcomeGRPO{} is less effective on Fine Cut, we evaluate every turn in the untrained model's validation rollouts from the second evaluation repetition using Gemini 3.8 Flash. The judge labels each turn as positive, neutral, or negative. We quantify rollout mixing as $2\min(n_+,n_-)/(n_++n_-)$, where $n_+$ and $n_-$ denote the numbers of positive and negative turns, respectively.

As shown in Table~\ref{tab:turn-quality}, Fine Cut has a mixing score of 0.651, whereas the other three task types range from 0.297 to 0.390. Trajectory-level credit assignment assigns the same advantage to every turn in a rollout and cannot distinguish between positive and negative turns.
The higher mixing on Fine Cut is consistent with greater ambiguity in assigning credit to individual turns.
This may partly explain the smaller gains from \outcomeGRPO{} on Fine Cut.

\begin{table}[htbp]
\caption{Turn-level quality of the untrained model's rollouts from the second
evaluation repetition on all 81 validation tasks, scored by the process-reward judge. }
\label{tab:turn-quality}
\begin{center}
\begin{tabular}{@{}lrrrr@{}}
\toprule
Task type & Positive & Neutral & Negative & Mixing \\
\midrule
CS & 55.3\% & 36.0\% & 8.7\% & 0.297 \\
NS & 42.7\% & 44.9\% & 12.4\% & 0.355 \\
RC & 44.5\% & 43.6\% & 11.9\% & 0.390 \\
FC & 16.7\% & 47.5\% & 35.8\% & \textbf{0.651} \\
\bottomrule
\end{tabular}
\end{center}
\end{table}

\subsection{GRPO+PRM: Reward Hacking and Mitigation}
\label{sec:analysis-hacking}
\label{sec:analysis-decay}
\paragraph{Reward hacking with a fixed \processRewardWeight{}.}
All task types except Fine Cut score lower than \outcomeGRPO{}, as shown in Table~\ref{tab:main-results}. Table~\ref{tab:interaction-behaviour} further shows a sharp decline in the model's ability to deliver a result, with 44.1\% of rollouts being truncated compared with only 0.3\% under \outcomeGRPO{}.
Late in training, the model continues calling tools to gather evidence but often stops before committing to a final result.

We further analyze the training dynamics of \GRPOPRMFixed{}. As shown in Figure~\ref{fig:teaser}(c), the process reward increases while the outcome reward declines after update 30. Appendix~\ref{app:training-dynamics} provides the full training curves.
As shown in Section~\ref{sec:model-behaviour}, process rewards encourage
desirable behaviors such as visual inspection, reasoning, and reflection.
One possible interpretation is that encouraging these behaviors helps improve
task scores early in training.
Later, continued optimization with a fixed process-reward weight may lead the
model to overuse these behaviors, consuming its interaction budget and reducing
its ability to complete the task.

\paragraph{\ProcessRewardWeight{} decay mitigates reward hacking.}

Observations in Figure~\ref{fig:teaser}(c) motivate us to decay the \processRewardWeight{} $\lambda$
to zero between updates 20 and 30.
This schedule retains process guidance early in training and shifts
later updates to the outcome reward alone.
Results in Table~\ref{tab:main-results} show that this approach is effective. On the first three task types, the model achieves performance comparable to \outcomeGRPO{}, with an average score of 0.568 versus 0.559. On Fine Cut, the method improves performance over the untrained model by 0.141, whereas \outcomeGRPO{} yields an improvement of only 0.050. As shown in Section~\ref{sec:model-behaviour}, decaying the
process-reward weight further encourages desirable behaviors,
such as longer reasoning and more frequent reflection,
compared with keeping the weight fixed.

\subsection{\OPSD{} Is Sensitive to \PrivilegedInformation{}}
\label{sec:analysis-opsd}
With \referenceAnswers{}, \OPSD{} scores 0.277, close to the untrained
model's 0.284. With \turnSuggestions{}, its score drops to 0.044
(Table~\ref{tab:main-results}).

\paragraph{\OPSD{} gradients depend on \privilegedInformation{}.}
The two variants share the same distillation loss, frozen teacher,
student model, and training hyperparameters.
Compared with \referenceAnswers{}, \turnSuggestions{} yield an
approximately 15-fold larger mean forward KL divergence and increase
the mean gradient norm from $0.084$ to $1.177$
(Table~\ref{tab:gradient-norm}).

\paragraph{The \OPSD{} signal can be diluted over a trajectory.}
With \referenceAnswers{}, the most divergent $5\%$ of turns account
for $47.4\%$ of the total KL but only $17.9\%$ of the tokens
(Table~\ref{tab:kl-concentration}).
This concentration suggests that averaging over valid assistant tokens
may dilute the contribution of the more divergent turns.
However, KL measures disagreement rather than the usefulness of
teacher guidance, and whether gradient magnitude or signal dilution
causes the observed outcomes remains to be tested.
Appendix~\ref{app:opsd-implementation} provides further analysis of
trajectory lengths, gradient norms, and KL concentration.

\section{Model Behavior}
\label{sec:model-behaviour}
Table~\ref{tab:interaction-behaviour} compares the behavior of the trained policies.
We use \outcomeGRPO{} as the baseline for the \GRPOPRM{} variants and the untrained model for \OPSD{}.
Reflection counts turns with an explicit self-correction marker; it does not assess whether a correction is valid.

\begin{table}[htbp]
\caption{Validation behavior across four repetitions of all 81 tasks.
Fixed-weight PRM has a truncation rate of 44.1\%; decaying the weight reduces
it to 6.2\%, with longer reasoning and more frequent reflection markers.}
\label{tab:interaction-behaviour}
\setlength{\tabcolsep}{2.5pt}
\renewcommand{\arraystretch}{1.12}
\begin{center}
\begin{tabular*}{\linewidth}{@{\extracolsep{\fill}}lrrrrrrr@{}}
\toprule
& \multicolumn{3}{c}{Interaction}
& \multicolumn{2}{c}{Visual inspection}
& \multicolumn{2}{c}{Reasoning} \\
\cmidrule(lr){2-4}\cmidrule(lr){5-6}\cmidrule(l){7-8}
Method & Turns & Tool calls & Truncated & Frames & Blind & Length & Reflection \\
\midrule
\baseModel{} & 41.7 & 61.4 & 1.5\% & 32.2 & 7.4\% & 644 & 17.7\% \\
\OutcomeGRPO{} & 45.1 & 52.2 & 0.3\% & 84.2 & 4.6\% & 618 & 19.0\% \\
\midrule
\GRPOPRMFixed{} & 82.4 & 165.8 & 44.1\% & 143.6 & 0.3\% & 761 & 24.8\% \\
\GRPOPRMDecayed{} & 56.9 & 99.5 & 6.2\% & 129.8 & 0.0\% & 1{,}035 & 36.0\% \\
\midrule
\OPSDReference{} & 41.1 & 62.5 & 1.5\% & 36.7 & 3.4\% & 657 & 17.8\% \\
\OPSDSuggestions{} & 11.8 & 19.3 & 0.0\% & 19.4 & 34.9\% & 59 & 0.7\% \\
\bottomrule
\end{tabular*}
\par\vspace{3pt}
\parbox{\linewidth}{Turns, tool calls, and frames are means per rollout; frames count returned images.
Truncated and Blind are percentages of rollouts stopped by a budget and receiving
no images, respectively. Length is reasoning characters per completed turn;
Reflection is the percentage of completed turns with an explicit self-correction marker.
See Appendix~\ref{app:behavior-measurement} for measurement details.}
\end{center}
\end{table}
\paragraph{Process rewards encourage desirable behaviors but do not ensure completion.}
Compared with \outcomeGRPO{}, \GRPOPRMFixed{} encourages more visual
inspection, longer reasoning, and more frequent self-correction markers
(Table~\ref{tab:interaction-behaviour}).
However, these changes accompany higher scores only on Fine Cut;
scores fall on the other three task types (Table~\ref{tab:main-results}).
The truncation rate also rises from 0.3\% to 44.1\%.
The value of these behaviors therefore depends on the task and on
whether the model can still complete it.

\paragraph{GRPO gains accompany greater visual inspection.}
Compared with the base model, \outcomeGRPO{} increases frames seen from
32.2 to 84.2, without increasing tool calls or reasoning length
(Table~\ref{tab:interaction-behaviour}).
Greater visual input may help the three shorter-horizon tasks,
consistent with their larger gains, whereas Fine Cut may require
more complex tool use and reasoning to coordinate a longer sequence
of editing operations.

\paragraph{Less process guidance mitigates hacking and elicits better behavioral patterns.}
Compared with \GRPOPRMFixed{}, \GRPOPRMDecayed{} makes fewer tool calls
while retaining substantial visual inspection, producing longer reasoning,
and showing more frequent self-correction markers
(Table~\ref{tab:interaction-behaviour}).
These changes accompany a reduction in truncation from $44.1\%$ to $6.2\%$
and an increase in overall task score from $0.334$ to $0.447$
(Table~\ref{tab:main-results}).
Compared with \outcomeGRPO{}, \GRPOPRMDecayed{} achieves a substantial
gain on Fine Cut, which may be associated with its longer reasoning
and more extensive visual inspection.

\paragraph{\OPSD{} behavior depends on \privilegedInformation{}.}
As shown in Table~\ref{tab:interaction-behaviour}, with
\referenceAnswers{}, \OPSD{} remains behaviorally close to the
untrained model, apart from a lower blind rate.
With \turnSuggestions{}, rollouts become much shorter, reasoning
length and reflection frequency drop sharply, and blind rollouts
become more frequent.
These behavioral changes accompany the collapse in task performance
(Table~\ref{tab:main-results}).

\FloatBarrier
\section{Conclusion}
\label{sec:conclusion}
Our experiments show when finer-grained credit helps and when it hurts.
Process guidance helps most on Fine Cut, while \outcomeGRPO{} remains
competitive on shorter-horizon tasks.
Process guidance can hurt when more visual inspection and reasoning
accompany poorer task completion. Decaying the process-reward
weight improves completion while retaining these behaviors.
\OPSD{} is sensitive to \privilegedInformation{}, showing that dense
supervision alone does not guarantee effective learning.
Our process-reward findings are based on GRPO+PRM with LLM-judged turn-level
rewards on movie-trailer editing tasks; whether they extend to untested methods
such as GiGPO~\citep{feng2025gigpo} and Turn-PPO~\citep{li2025turnppo} remains open.
For \OPSD{}, we identify failure patterns without establishing effective
remedies. How to improve teacher guidance and training stability requires
further study.

\clearpage
\bibliographystyle{ims-arxiv}
\bibliography{references}

\clearpage
\appendix
\section*{Contents of the Appendix}
\startcontents[appendix]
\begingroup
  \hypersetup{linkcolor=black}
  \printcontents[appendix]{app}{1}[2]{%
    \setlength{\parskip}{0pt}%
    \contentsmargin{20pt}%
  }
\endgroup
\clearpage
\section{Task Construction Details}
\label{app:task-construction}

\begin{figure}[htbp]
\centering
\includegraphics[width=\linewidth,height=0.80\textheight,keepaspectratio]{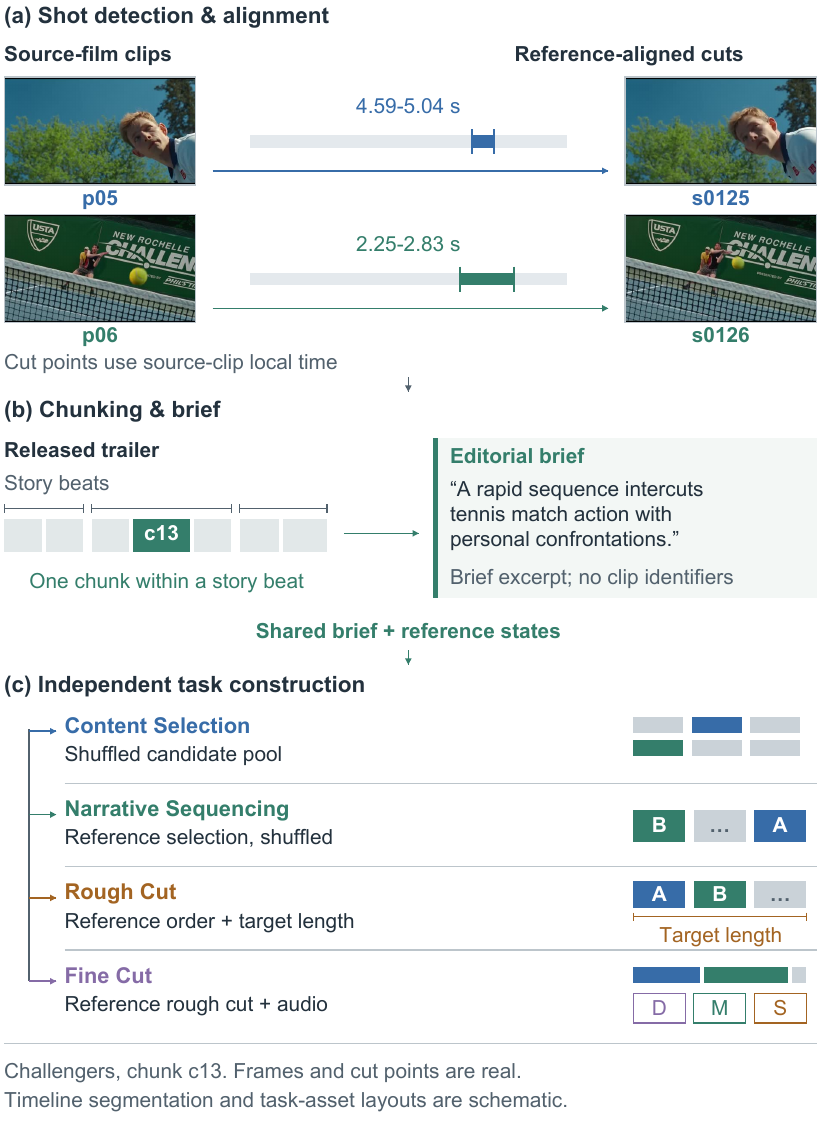}
\caption{Data generation from a film--trailer pair. (a) Two real source clips and their
reference-aligned cuts illustrate shot matching. (b) Each local chunk stays within a story
beat and receives a clip-independent brief. (c) Four tasks share the brief and initialize
from reference states, so errors in sampled outputs cannot propagate between tasks.
Timeline and asset layouts are schematic. D/M/S denote dialogue, music, and production sound.}
\label{fig:pipeline}
\end{figure}

Figure~\ref{fig:pipeline} gives an overview of the data generation pipeline.

\subsection{Corpus and Summary Statistics}

Let
\begin{equation}
    \mathcal{D}_{\mathrm{edit}}=\{(F_i,T_i)\}_{i=1}^{N},
\end{equation}
where $F_i$ is a source film and $T_i$ is its corresponding trailer. The corpus
contains $N=49$ pairs, comprising 101.80 hours of source video and 2.09 hours of
trailers. Instantiating four task types for every retained chunk produces
an initial pool of 2,352 tasks, before task-level quality filtering and sampling.
The final experimental dataset contains 797 tasks
(Section~\ref{sec:trailer-data-filtering}).

\subsection{Media Processing and Chunking}

Task construction converts each long film and short trailer into localized editing
problems. We normalize media geometry and create analysis proxies, then detect shot
boundaries in both videos. A coarse-to-fine visual matching pipeline aligns trailer
shots to candidate source-film moments and rejects low-confidence matches. Graphic
cards, black frames, flashes, and other featureless intervals are separated from
usable footage. Speech timestamps, music activity, and picture--sound relations are
extracted only for constructing fine-cut tasks and hidden evaluation evidence; the
finished trailer mix and its alignments are never exposed to the policy.

A semantic pass groups the trailer into broad story beats and then into local chunks
that do not cross beat boundaries. The construction produces 622 raw chunks. We
retain 588 chunks containing at least three usable picture clips and two source-film
anchors. The source films contain 88,669 detected shots and the trailers contain
5,978 detected shots.

\subsection{Construction by Task Type}

The four tasks for a chunk are constructed independently. Narrative Sequencing uses
the reference selection, Rough Cut uses the reference order, and Fine Cut starts from
a reference rough cut rather than a sampled upstream output. This prevents one
task type's errors from dominating another task type's training signal. The corpus
therefore evaluates stagewise editing decisions; composing all task types into a
fully end-to-end trailer editor is a separate setting.

Content Selection presents complete source shots rather than pre-trimmed positives
and mixes them with neighboring and duration-matched distractors. In the final dataset, each task contains
21--36 candidates and 3--14 positives. Narrative Sequencing independently renames
and shuffles the selected clips. Overlapping neighboring source windows are separated
where possible so that the task cannot be solved by matching duplicate boundary
frames. Rough Cut supplies the reference order but requires new in- and out-points
for every clip. Matching ambiguity is retained as a timing tolerance, avoiding
frame-exact penalties when equivalent boundaries exist.

Fine Cut begins from a stable rough cut and exposes production-sound handles, uncut
dialogue candidates, an unaligned music window, and join metadata. The agent selects
and trims dialogue, places music, chooses picture and sound transitions, and renders
the result. Trailer-only picture is muted so that released dialogue and music choices
cannot be recovered from its audio. The structured edit and rendered preview are
evaluated together.

\subsection{Task Inputs, Outputs, and Requirements}
\label{app:task-interfaces}

Each task package provides a local editing brief, the assets under
\texttt{initial\_state/}, and a public \texttt{manifest.json} describing those
assets. The manifest identifies files and their local time coordinates; it is not
an answer key. Hidden selections, reference orders, source--trailer alignments,
and grading references are kept outside the policy-visible package. The four
interfaces expose different decisions, as detailed below. Their scores use
different criteria and should not be interpreted as a common difficulty scale.

\paragraph{Content Selection (CS): choose an unordered subset.}
The input is a written description of the local trailer stretch and a pool of
complete, silent source shots, with candidate IDs, file names, and durations in
the manifest. Positive shots are mixed with distractors from the same film.
Candidate numbering follows a shuffle and does not indicate relevance.
The agent must screen the pool and decide both which clips to keep and how many
to keep. It writes \texttt{output/selection.json}, containing a
\texttt{selected} array of candidate IDs. Each selected ID is listed once;
the order of IDs has no meaning. The task does not ask the agent to rank the
candidates, order the selected shots, choose temporal boundaries, mix audio,
or render a video.

The score is the set-selection F1 against the hidden selection: precision is the
fraction of selected valid candidates that belong to the reference set, and
recall is the fraction of reference candidates selected. An empty selection
receives zero. The reference selection size is not supplied and is not a
submission constraint; selecting too many clips is penalized through precision
and selecting too few through recall. This is a continuous F1 score, not a
binary task-success rate. The task grader additionally reports missed and
incorrect clips; its diagnostic pass flag is separate from the reported score.

\paragraph{Narrative Sequencing (NS): order the supplied shots.}
The input contains the reference-selected shots, independently shuffled and
renamed, together with a description of the desired narrative stretch.
Unlike the silent CS pool, the NS clips include production sound. The agent can
use action continuity, dialogue, eyelines, movement, shot size, scene changes,
and the brief to recover an ordering. It writes \texttt{output/order.json},
whose \texttt{order} array lists every supplied ID exactly once, from first to
last. Dropping a shot, duplicating an ID, or introducing a new ID is invalid.
The agent does not choose a subset, trim shots, or alter the sound mix.

The deterministic score combines three complementary properties of the
permutation: one minus normalized positional displacement, the normalized
longest increasing subsequence, and the fraction of directed reference-adjacent
pairs retained. These three factors are multiplied. Thus, preserving some
pairwise order alone does not ensure that the sequence preserves the reference
joins. The \emph{Heretic} chunk~5 grader used in our experiments accepts four explicitly
enumerated orders and takes the best match among them; this is a task-specific
repair rather than a general acceptance of arbitrary alternative orders.

\paragraph{Rough Cut (RC): choose an interval within every ordered clip.}
The input supplies the selected clips in their reference order, their available
durations, a local editing brief, and a target total duration. The agent writes
\texttt{output/cuts.json}. Each object in its \texttt{cuts} array contains
\texttt{clip}, \texttt{in\_s}, and \texttt{out\_s}. Times are measured from
the beginning of the named source clip, rather than on a global film or output
timeline. All supplied clips must appear exactly once and in the supplied
order. Every kept interval must have a nonnegative start, an end after its
start, and boundaries within the available media. The sum of the kept
durations must stay near the supplied target. The agent chooses where an action
begins and ends and how much time each shot receives; it does not reselect or
reorder the material, add dialogue, or redesign the mix. The grading pipeline
renders the submitted cuts for evaluation.

The final RC score combines a deterministic timing component (weight 0.4),
reference-informed judgment (0.4), and judgment without the reference (0.2).
The deterministic component is the fraction of scorable shots for which both
the in-point error and the kept-duration error are within their reference
tolerances. In-point tolerances are specific to each shot, reflecting matching
ambiguity; they are not a universal frame-exact threshold. Shots with no
source-film counterpart remain part of the required sequence but are excluded
from this timing-component denominator. Submission validity and the full
judging rubrics are detailed in Appendix~\ref{app:experimental-details}.

\paragraph{Fine Cut (FC): finish sound and transitions on a fixed picture edit.}
The input is a reference rough cut with fixed shot order and picture timing.
The manifest supplies shot durations, the times and identities of joins,
available production-sound handles, an index of uncut dialogue takes, and an
unaligned music window. These are candidate materials, not an assembled
reference mix. Dialogue takes can contain surrounding speech, and some takes
do not belong in the requested stretch. A shot made for the trailer can arrive
without production sound; the prompt and manifest identify such cases.

The agent must deliver both \texttt{output/edit.json} and
\texttt{output/preview.mp4}. The structured edit records the following choices:
\begin{itemize}
    \item \textbf{Dialogue:} which supplied takes to use, the source-local
    \texttt{in\_s}/\texttt{out\_s} interval in each take, the output-timeline
    placement \texttt{at\_s}, and any gain adjustment. Unneeded takes may be
    omitted; a take may be reused when the edit calls for it.
    \item \textbf{Production sound:} whether to retain or mute a shot's own
    sound and, when a handle is used, its source interval and timeline
    placement. Silent shots cannot supply production sound that does not exist.
    \item \textbf{Music:} the offset within the supplied cue that aligns
    with the start of the edit, and the resulting audible entry and balance.
    The agent must determine the alignment from the material.
    \item \textbf{Every join:} a picture choice of \texttt{cut},
    \texttt{dip\_to\_black}, \texttt{dissolve}, or \texttt{other}, and a sound
    choice of \texttt{hard}, \texttt{carries\_over}, \texttt{prelap}, or
    \texttt{under}. These respectively distinguish a sound break, an L-cut,
    a J-cut, and a sound bridge. Every supplied join must be specified once;
    there is no quota requiring bridges or decorative transitions.
    \item \textbf{Cards, when required by the task:} text and appearance for
    supplied card placeholders. These are task-dependent, not a requirement to
    add a card to every stretch.
\end{itemize}
The picture shots and their established timing remain fixed, subject to the
requested join treatments. The rendered preview must make the declared choices
audible and visible and stay within the duration allowance used by the
submission validator. Merely declaring a line in JSON does not establish that
it can be heard in the final mix. The FC score combines judgment without the
reference (weight 0.6) and reference-informed judgment (0.4); there is no
deterministic exact-match component. The reference is one realized solution,
and the rubrics evaluate task-appropriate alternatives through their actual
audiovisual result.

\subsection{Audit, Repair and Sampling}

We audit every constructed task with Gemini 3.8 Flash, giving it the task description, the
available tools, the video and audio assets, and the hidden grading references. The audit
checks whether the brief is supported by the supplied media and whether the grading rules
evaluate decisions that can be inferred from the public input. For Content Selection and
Narrative Sequencing it checks whether visual and narrative cues support the target
selection or order. For Rough Cut and Fine Cut it checks the feasibility of the timing,
dialogue and audiovisual requirements, and it allows several reasonable edits.

Tasks with contradictory briefs, missing or corrupted assets, answer leakage, or
unsupported reference constraints are repaired or dropped. A repair makes targeted changes
to prompts, references, graders or media, followed by file-integrity, decoding and grader
checks as applicable. Unresolved tasks and incomplete audit responses leave the candidate
pool. Review coverage varies, and some candidates receive additional independent-context
reviews while others receive a single audit or repair-author judgment.

Filtering and repair leave 925 candidates. We draw 797 of them by fixed-seed sampling
stratified by split, task type and film, which preserves film coverage within each type
and uses no policy reward scores. The 128 unselected candidates remain valid and are set
aside only to fix the dataset size and the type proportions.

\subsection{Split and Anti-Leakage Controls}

The initial construction uses a film-level split: 43 films and 2,076 tasks in the
training pool, and 6 films and 276 tasks in the validation pool. Filtering and
sampling retain this film partition, yielding 716 training and 81 validation
tasks. The 43 training films are \emph{28 Years Later - The Bone Temple}, \emph{A Complete Unknown}, \emph{Arthur the King}, \emph{Avatar - The Way of Water}, \emph{Ballerina}, \emph{Beetlejuice Beetlejuice}, \emph{Bugonia}, \emph{Challengers}, \emph{Crime 101}, \emph{Dune - Part Two}, \emph{Evil Dead Burn}, \emph{Furious 7}, \emph{His Three Daughters}, \emph{Hit Man}, \emph{Horizon - An American Saga - Chapter 1}, \emph{How to Train Your Dragon}, \emph{John Wick - Chapter 4}, \emph{Land of Bad}, \emph{Lee}, \emph{M3GAN 2.0}, \emph{Maria}, \emph{May December}, \emph{Mercy}, \emph{Migration}, \emph{Mission Impossible - The Final Reckoning}, \emph{Mufasa - The Lion King}, \emph{Oppenheimer}, \emph{Ordinary Angels}, \emph{Ricky Stanicky}, \emph{Send Help}, \emph{Smile 2}, \emph{Song Sung Blue}, \emph{The Bikeriders}, \emph{The Hunger Games - The Ballad of Songbirds \& Snakes}, \emph{The Iron Claw}, \emph{The Marvels}, \emph{The Rip}, \emph{The Wild Robot}, \emph{Transformers One}, \emph{Warfare}, \emph{White Bird}, \emph{Young Woman and the Sea}, \emph{Zootopia 2}. The 6 validation films are \emph{Deadpool \& Wolverine}, \emph{Heretic}, \emph{Sinners}, \emph{The Drama}, \emph{They Will Kill You} and \emph{Whiplash}. Candidate identifiers are
independently randomized across task types, competing clips share a normalized
encoding path, and answer records and reference alignments remain outside the
policy-visible task package.

Table~\ref{tab:validation-film-coverage} lists the final validation coverage.
Filtering does not require every held-out film to contribute a task of every
type; for example, no Content Selection task from \emph{Deadpool \& Wolverine}
is retained in this split.
\begin{table}[htbp]
\caption{Final validation tasks by film and task type. These counts describe
the 81 tasks, before repeated evaluation.}
\label{tab:validation-film-coverage}
\begin{center}
\begin{tabular}{lrrrr}
\toprule
Film & CS & NS & RC & FC \\
\midrule
Heretic & 4 & 2 & 5 & 4 \\
Sinners & 4 & 2 & 4 & 4 \\
The Drama & 2 & 3 & 3 & 5 \\
They Will Kill You & 1 & 2 & 3 & 4 \\
Whiplash & 3 & 3 & 7 & 4 \\
Deadpool \& Wolverine & 0 & 3 & 4 & 5 \\
\midrule
Total & 14 & 15 & 26 & 26 \\
\bottomrule
\end{tabular}
\end{center}
\end{table}

\subsection{Example Tasks}
\label{app:example-tasks}

We expand the two worked task examples, Content Selection and Fine Cut, below
and then provide additional Narrative Sequencing and Rough Cut examples.
All four are actual validation tasks from the dataset used in our experiments. The task-specific
prompts are reproduced in full, with only typesetting and JSON whitespace changed;
the shared environment and tool instructions are omitted. File names in the
prompts are relative to the task directory or its input asset directory.
The JSON blocks are the format illustrations supplied in those prompts, not
reference answers or completed submissions. In particular, the Rough Cut block
lists only two placeholder intervals and the Fine Cut block lists only four of
seven joins. An actual submission must satisfy the complete task requirements.
The manifest details preceding each prompt report the actual supplied assets.

\paragraph{Content Selection: \emph{Heretic}, chunk 1.}
Dataset ID: \texttt{s\_1\_heretic\_c1}; task-local ID:
\texttt{heretic\_s1\_c1}. The public manifest contains 30 candidates
(\texttt{c01}--\texttt{c32}, excluding \texttt{c13} and \texttt{c15}), each
with a path under \texttt{initial\_state/clips/} and its duration.
The hidden reference contains 11 selected shots; this number is not supplied to
the agent. The task asks for selection only.

\begin{quote}
\normalsize
\textbf{Task --- pick the clips for this stretch.}\par
You are the assistant editor on the \emph{Heretic} trailer. There are \textbf{30 candidate clips} under \texttt{initial\_state/clips/}.

This stretch of the trailer covers:

\emph{Two young women walk up a wooded driveway where their bicycles are parked and are greeted at the front door as the homeowner opens it. In the living room, a vinyl record spins on a turntable and the homeowner presents a closed copy of the Book of Mormon on the table. The conversation is warm with shared smiles and laughter across the room, concluding with him leaning forward at his desk speaking enthusiastically, followed by a close-up of the young woman on the couch reacting with an amused, questioning head tilt.}

It sits inside the beat \textbf{The Doorstep}.

\textbf{What to do.}\par
Screen every candidate and pick the ones this stretch actually uses --- the shots that cut together into the description above. The rest are real footage from the same film that the trailer did not use here; most are neighbouring shots from the same scenes.

\begin{itemize}
\item Order does not matter here. This stage is only about which clips.
\item The count is not given. Decide it yourself; over-picking costs as much as
under-picking.
\item Every candidate is a \textbf{complete shot at its own boundaries} --- silent, identically
transcoded, numbered after a shuffle. \textbf{Judge from the picture.}
\end{itemize}
\textbf{Deliverable.}\par\nopagebreak
Write \texttt{output/selection.json}:

\end{quote}

\noindent\begin{minipage}{\linewidth}
\normalsize
\begin{verbatim}
{"selected": ["c03", "c07", "c19"]}
\end{verbatim}
\end{minipage}

\begin{quote}
Just the ids you picked, in any order.

\end{quote}

\paragraph{Fine Cut: \emph{Deadpool \& Wolverine}, chunk 3.}
Dataset ID: \texttt{s\_4e\_deadpoolwolverine\_c3}; task-local ID:
\texttt{deadpoolwolverine\_s4e\_c3}. The manifest fixes a 12.867-second rough
cut at 23.976024 frames per second, with shots \texttt{p01}--\texttt{p08} and
seven joins. Their successive durations are 1.272, 3.212, 1.835, 0.876, 1.001,
1.376, 1.377, and 1.918 seconds. The join times are 1.272, 4.484, 6.319, 7.195,
8.196, 9.572, and 10.949 seconds on the rough-cut timeline. Shot \texttt{p06}
is trailer-only and has no production sound; the other seven shots have sound
handles extending four seconds on each side. Five dialogue candidates
(\texttt{d01}--\texttt{d05}) and the music window complete the input.

\begin{quote}
\normalsize
\setlength{\emergencystretch}{4em}
\textbf{Fine cut --- deadpoolwolverine / c3.}\par
You have a rough cut: 8 shots in order, each already trimmed to the frame, joined with 7 straight cuts. The pictures and their timing are settled and \textbf{are not yours to change}.

What is not settled is everything you can hear, and what happens at the joins.

This stretch covers:

\emph{Wide of the wasteland and Wolverine's yellow-suit reveal in blinding light, cut into the birthday party for "I'm about to lose everything I've ever cared about," then back to the wasteland for the two of them going at it --- "not my problem" / "is that what you said when your world went to shit?"}

\textbf{What you have.}\par
\begin{itemize}
\item \texttt{shots/pNN.mp4}: the rough cut's shots. They carry the \textbf{film's own production sound} --- whatever the microphone picked up on set.
\item \texttt{handles/pNN.m4a}: the same sound, but running 4 seconds before each shot's first frame and 4 seconds past its last. \texttt{manifest.json} gives \texttt{handle.picture\_in\_s} --- where the picture starts inside that file.
\item \texttt{takes/dNN.m4a}: \textbf{5 pieces of dialogue, lifted from the film uncut.} \texttt{manifest.json} lists every stretch of speech in each file (\texttt{speech\_spans\_s}) --- the line that was looked for, and whatever else happens to sit either side of it. All of it is yours to use. Not all of these pieces belong in this stretch.
\item \texttt{music.wav}: the trailer's music. The window runs wider than this stretch at both ends --- how much wider is not stated --- and it is \textbf{not lined up}. Where the music comes in is one of your decisions.
\item \texttt{manifest.json}: shots, joins with their exact times and frames, dialogue index, music window.
\end{itemize}
\textbf{The work.}\par
\textbf{Sound.} The rough cut plays each shot's own production sound, cut off hard at both edges. That is not a trailer. Decide what is heard:

\begin{itemize}
\item Which of the 5 lines belong in this stretch, and which do not. Some are here to be left out.
\item Where each line you keep starts --- under which shot, and how much of it runs before or after the picture changes.
\item How much of a take to use. They run long on purpose; the trailer's version of a line is usually shorter than the film's.
\item Where the production sound is worth keeping, and where it should give way.
\item Where the music enters, and how it sits against the pictures.
\end{itemize}
\textbf{Silent shots.} p06 was made for the trailer rather than shot for the film, so it has no production sound at all --- the picture arrives silent. What plays under it is entirely your call.

\textbf{Picture.} Most joins want a straight cut and nothing else --- this is the common case, and adding something everywhere is the most reliable way to make a trailer worse. Where a join wants more, the moves are: a \textbf{dip to black} (how long it holds, and whether either side ramps, is part of the call), a \textbf{dissolve}, or something faster --- a whip pan, a flash frame.

None of this is a checklist. Which of these a join wants --- or whether it wants any --- is the question being asked.

\textbf{Deliverable.}\par\nopagebreak
Write \texttt{output/edit.json}:

\end{quote}

\noindent\begin{minipage}{\linewidth}
\normalsize
\begin{verbatim}
{
  "music": {"starts_at_s": 3.20,
    "note": "file time that lands on 0.0 of the cut"},
  "dialogue": [
    {"take": "d03", "in_s": 1.40, "out_s": 3.10,
     "at_s": 0.00, "gain_db": 14,
     "why":
"carries into the second shot before the picture turns"
    }
  ],
  "joins": [
    {"join": 1, "picture": "cut", "sound": "hard"},
    {"join": 2, "picture": "dip_to_black",
      "black_hold_s": 0.30,
      "fade_in_s": 0.35, "sound": "under"},
    {"join": 3, "picture": "cut",
      "sound": "carries_over",
      "note": "line d03 runs on"},
    {"join": 4, "picture": "cut", "sound": "under",
      "note": "music holds it"}
  ],
  "production_sound": [
    {"shot": "p04", "keep": true, "gain_db": -6}],
  "cards": [{"shot": "p04",
    "text": "FROM THE DIRECTOR OF",
    "look": "white type, centred, on black",
    "in": "cut", "out": "dip"}]
}
\end{verbatim}
\end{minipage}

\begin{quote}
\texttt{take} / \texttt{in\_s} / \texttt{out\_s} are which line and which part of it; \texttt{at\_s} is where it lands on this stretch's timeline, where 0.0 is the first frame. \texttt{gain\_db} is how far you lift or drop it.

\textbf{On levels.} The music was mixed against a different set of lines than the ones you have, so it does not step aside where you need it to. Measured across the corpus, a take sits about 12 dB \emph{under} the music bed, and dialogue that reads clearly usually sits 6--12 dB \emph{over} it. Whatever you lay in will be buried unless you say otherwise --- \texttt{gain\_db} on the line, \texttt{gain\_db} on the production sound, or both.

On each join:

\begin{itemize}
\item \texttt{picture} --- one of \texttt{cut} / \texttt{dip\_to\_black} / \texttt{dissolve} / \texttt{other}
\item \texttt{sound} --- one of these four, and give it on \textbf{every} join:
\begin{itemize}
\item \texttt{hard} --- the sound breaks where the picture breaks
\item \texttt{carries\_over} --- something belonging to the leaving shot runs past it. An \textbf{L-cut}.
\item \texttt{prelap} --- something belonging to the arriving shot starts early. A \textbf{J-cut}.
\item \texttt{under} --- neither side's own sound matters here; what carries the join is laid on top. A \textbf{sound bridge} --- usually the music bed, sometimes a line from elsewhere running across both shots.
\end{itemize}
\end{itemize}
There are \textbf{7 joins}, numbered 1 to 7 in order. Give all of them.

\textbf{Also render it.} Build the result as \texttt{output/preview.mp4} --- this stretch's picture with your mix under it. It runs the length of the rough cut plus whatever black you chose to add. \texttt{ffmpeg} is here. The JSON says what you decided; the render is where anyone can hear whether it worked. A line you placed but cannot hear in the render is a line you did not place.

\textbf{How this is judged.}\par
Not against a single correct answer. The trailer that was actually cut from this material is one solution; a different set of choices can be as good or better. What is looked at is whether the choices hold together: does a line land where the picture supports it, does the music arrive somewhere that means something, is the restraint at the plain joins deliberate rather than absent.

\end{quote}

\paragraph{Narrative Sequencing: \emph{Heretic}, chunk 5.}
Dataset ID: \texttt{s\_2\_heretic\_c5}; task-local ID:
\texttt{heretic\_s2\_c5}. The manifest supplies seven shuffled clips with
production sound, \texttt{k01}--\texttt{k07}. Their durations, in ID order,
are 9.215, 9.215, 8.881, 7.963, 5.752, 7.087, and 5.961 seconds. The output must
be a permutation of these seven IDs; the IDs are not narrative positions.

\begin{quote}
\normalsize
\textbf{Task --- put this stretch in order.}\par
You are the assistant editor on the \emph{Heretic} trailer. The \textbf{7 shots} this stretch uses have already been chosen and sit in \texttt{initial\_state/clips/} --- but they are \textbf{shuffled}.

This stretch covers:

\emph{After the second press card, Mr. Reed is at his workbench assembling the architectural model of the house, followed by a close-up of his hands trimming a figurine with a craft blade. Next, Mr. Reed stands lurking in the shadowy hallway. Cut back to the two women confronting the test at the doors: Barnes crying in despair against the doorframe, and Paxton approaching the door marked "DISBELIEF". Finally, reaction shots of escalating dread: a tear falling down Paxton's cheek, tightening to an extreme close-up of her throat swallowing as Mr. Reed describes the physical grip of fear.}

It sits inside the beat \textbf{Being Studied}.

\textbf{What to do.}\par
Put the 7 shots into the order the trailer used.

Use all of them, each exactly once. The ids were assigned after a shuffle and \textbf{carry no ordering information}.

Judge from picture and sound: how actions run on from one shot to the next, dialogue continuity, where eyelines and movement point, how shot sizes step, where the location changes, and the narrative progression in the description above.

\Needspace{8\baselineskip}
\textbf{Deliverable.}\par\nopagebreak
Write \texttt{output/order.json}:

\end{quote}

\noindent\begin{minipage}{\linewidth}
\normalsize
\begin{verbatim}
{"order": ["k03", "k01", "k05", "k07",
           "k06", "k02", "k04"]}
\end{verbatim}
\end{minipage}

\begin{quote}
Every id, first to last.

\end{quote}

\paragraph{Rough Cut: \emph{Heretic}, chunk 2.}
Dataset ID: \texttt{s\_3\_heretic\_c2}; task-local ID:
\texttt{heretic\_s3\_c2}. The manifest supplies 11 ordered source clips,
\texttt{p01}--\texttt{p11}, with durations 3.698, 2.496, 3.674, 9.257,
9.840, 9.855, 2.339, 12.431, 18.185, 20.976, and 19.976 seconds. The exact
manifest target is 15.182 seconds, rounded to 15.18 seconds in the prompt.
Every clip needs its own interval. The numeric intervals in the prompt's JSON
block illustrate field syntax only and must be replaced by intervals within
these actual source durations.

\begin{quote}
\normalsize
\textbf{Task --- rough-cut this stretch.}\par
You are the assistant editor on the \emph{Heretic} trailer. The shots for this stretch are chosen and ordered already, in \texttt{initial\_state/clips/} --- \textbf{11 of them}, \texttt{p01} first.

Your job is to \textbf{set an in point and an out point on each} --- where to start using that shot and where to leave it. Most run longer than the trailer's use of them.

This stretch covers:

\emph{A run of dark close-ups --- locks, a dial, ceiling lights from below --- establishes the house as a maze. The two are led into the peaked-roof room with two doors side by side, and he tells them leaving means going out the back of his house.}

\textbf{Target total: 15.18 seconds} across the 11 shots.

\textbf{What to do.}\par
For each shot, decide which part of it the trailer would use:

\begin{itemize}
\item where the action reads strongest --- where to come in, where to leave
\item whether to cut away mid-gesture or let it finish
\item the lengths should add up to close to the target
\end{itemize}
\textbf{Deliverable.}\par\nopagebreak
Write \texttt{output/cuts.json}:

\end{quote}

\noindent\begin{minipage}{\linewidth}
\normalsize
\begin{verbatim}
{"cuts": [
  {"clip": "p01", "in_s": 12.34, "out_s": 13.90},
  {"clip": "p02", "in_s": 4.05,  "out_s": 5.20}
]}
\end{verbatim}
\end{minipage}

\begin{quote}
\texttt{in\_s} / \texttt{out\_s} are seconds \textbf{within that clip} (the clip starts at 0). Give all 11, in order.

\end{quote}

\clearpage
\section{Editing Environment Details}
\label{app:editing-environment}
\suppressfloats[t]

\subsection{Episode State and Tools}

Each task runs in a fresh filesystem workspace. The loader stages the task prompt,
a manifest, and only the media visible for the current task type. The public
manifest identifies the supplied assets and the timing information appropriate
to that task. Prepared frames and observation tools support Content Selection,
Narrative Sequencing, and Rough Cut. Fine Cut additionally requires manipulation
of the supplied picture, dialogue, production-sound and music assets to build
an audiovisual timeline and render a preview. Hidden answers, matching records,
rubrics, and reference decisions remain in a separate evaluator-owned directory.

The policy may maintain notes, compare candidates, write timeline scripts, render
intermediate results, inspect them, and revise its output. A turn is one
policy generation request; it may contain several tool calls, so turn count
and tool-call count differ. The episode terminates when the policy produces
a final response or exhausts a turn, context, or output budget. The common
turn limit is 100. Termination by a final response does not establish that a
valid deliverable exists or that the task has been completed successfully.

\subsection{Tools}
\label{app:tools}

Beyond the harness's own shell, which provides FFmpeg, ffprobe and Python for local
cutting, joining, mixing and probing, the policy calls eight media tools over MCP.

\begin{table}[tbp]
\caption{Media tools available to the policy. None of them exposes a task label, a
hidden reference decision or an editing-specific oracle.}
\label{tab:tools}
\begin{center}
\begin{tabularx}{\linewidth}{@{}l>{\raggedright\arraybackslash}X@{}}
\toprule
Tool & What it returns \\
\midrule
\texttt{view\_image} & Prepared or policy-generated still images selected by path. \\
\texttt{inspect\_video} & Sampled video frames selected by timestamps or an interval;
returns actual decoded times rather than continuous playback. \\
\texttt{inspect\_frames} & Frame-index or time-based inspection near a cut, optionally
cropped; returns a sheet or individual images. \\
\texttt{contact\_sheet} & A numbered sheet of up to 16 requested source images or frames. \\
\texttt{inspect\_audio} & Duration, level, silence and an offline English transcript
for a requested audio window. \\
\texttt{analyze\_audio} & Loudness, true peak, channel-level measurements, silence,
and onset or pulse candidates for an audio window. \\
\texttt{validate\_media} & Technical checks on a declared edit or rendered file,
including timing, stream/decode information and schema constraints. \\
\texttt{listen\_music} & Pre-computed instrument, mood and energy descriptions
for a selected music window. \\
\bottomrule
\end{tabularx}
\end{center}
\end{table}

The policy interface supports text and images. The audio tools return analysis
text rather than an audio waveform for the model to listen to. A transcript may drop words, so
the absence of a transcript does not mean the absence of speech. The music tags come from
a read-only store built before training, not from an online model. Audio onsets
and periodicity estimates are candidates for editing decisions, not guaranteed
musical beats. The technical validator does not provide the hidden outcome
score or certify artistic quality. Thus observation and validation tools can
reduce uncertainty while still leaving the editing decision to the policy.

\subsection{Harness}

Episodes run in the official OpenCode harness at version 1.15.5, commit
\texttt{d7a6e1da}. We do not modify its binary or its upstream source. The media tools
above are served over MCP, and an image-registration plugin on the \texttt{tool.execute.after} hook
intercepts both native image attachments and MCP image content, normalizing and
registering each image before it reaches the conversation history. The
recorded configuration also disables automatic compaction and pruning and
retains the policy's original generation inputs for replay.

The plugin exists because the harness rearranges tool images into fresh user messages and
may compress or trim history. Registering images ahead of that step keeps the request the
inference backend receives identical to the one a training replay reconstructs, so the
policy is never optimized against pixels it did not see. Images that arrive by URL, in an
unknown binary modality, or without registration are refused rather than silently
downscaled or dropped.

\subsection{Deliverable Validation}

At termination, the environment collects only the named deliverables:
\texttt{selection.json} for Content Selection, \texttt{order.json} for
Narrative Sequencing, \texttt{cuts.json} for Rough Cut, and
\texttt{edit.json} together with \texttt{preview.mp4} for Fine Cut.
The submitted text response is not a substitute for these artifacts.
JSON validation checks types, identifiers, finite time values and the
schema required by the task. Missing or malformed deliverables and violated
submission constraints receive zero outcome reward. Infrastructure failures
and unavailable grades are tracked separately rather than automatically
counted as policy failures.

\paragraph{Rough Cut checks.}
Every supplied clip must occur exactly once and in the supplied order.
Each retained interval must have a nonnegative in-point, an out-point after
the in-point, and endpoints within the corresponding media. If the manifest
specifies a target duration $T$, the sum of all retained intervals must be
within $\max(1,\min(3,0.1T))$ seconds of $T$. This common delivery rule
supersedes older total-duration tolerance text in the raw task prompt; it
does not permit endpoints outside the source media. The original task-specific
reference tolerances still define the separate deterministic quality metric
(Appendix~\ref{app:outcome-scoring}).

\paragraph{Fine Cut checks.}
The edit description must cover every supplied join exactly once. Picture
join types are \texttt{cut}, \texttt{dip\_to\_black}, \texttt{dissolve} or
\texttt{other}; sound relationships are \texttt{hard}, \texttt{under},
\texttt{prelap} or \texttt{carries\_over}. Dialogue entries must reference
available takes with valid trim intervals and placement times. Production
sound must reference an existing source; explicit timing requires a supplied
handle and all timing fields. Required title cards must be declared with
nonempty text. A declared music offset must lie inside the supplied cue.
The preview must be probeable, contain a video stream, and meet the same
total-duration tolerance relative to the locked picture sequence. These
structural checks do not by themselves prove that dialogue is intelligible,
a declared transition appears in the render, or the mix satisfies the brief;
the outcome rubrics judge the actual delivered audiovisual result.

\subsection{Isolation and Telemetry}

Every trajectory executes in a disposable sandbox without network access.
Concurrent rollouts of the same task receive private workspaces, so one sample
cannot inspect or overwrite another sample's output. The environment logs turns,
tool calls, images viewed, validation errors, rendering status, and reward
components. These records support the later analysis of training cost, failure
modes, and credit assignment without enlarging the policy-visible state.

The same public task, observation tools and deliverable contract are used
for student training rollouts and evaluation. Reward computations and
OPSD teacher-only inputs are performed outside the policy sandbox. They do
not enlarge the student's public state or expose validation answers to it.

\FloatBarrier
\section{Training and Evaluation Details}
\label{app:experimental-details}

\FloatBarrier
\subsection{Training Configuration and Optimization}
\label{app:optimization}

\paragraph{Training configuration.}
Training uses asynchronous policy updates in SkyRL, with FSDP sharding and
vLLM inference engines. Table~\ref{tab:shared-training-config}
lists the settings shared by every training variant, and
Table~\ref{tab:per-arm-config} lists the settings that differ. The periodic validation pass uses one rollout per task to monitor progress.
Final reported scores use repeated evaluation of the selected final checkpoint;
where available, its recorded validation pass supplies the first repetition and
three additional passes complete avg@4. These repetitions vary rollout sampling,
not the training seed.

\begin{table}[htbp]
\caption{Training settings shared by all training variants.}
\label{tab:shared-training-config}
\begin{center}
\begin{tabularx}{\linewidth}{@{}l>{\raggedright\arraybackslash}X@{}}
\toprule
Setting & Value \\
\midrule
Tasks per update & 16 \\
Optimizer & AdamW, constant learning rate $1\times10^{-6}$ \\
Micro-batch per GPU & 2 \\
Training updates & 50 \\
Clipping thresholds $(\delta_{\mathrm{low}},\delta_{\mathrm{high}})$ & $(0.2,\,0.28)$ \\
Loss reduction & token mean \\
Additional batch-level advantage normalization & disabled \\
Rollout sampling & temperature 1.0, top-$p$ 1.0, top-$K$ $-1$ \\
Turn budget & 100 \\
Trainer prompt-length setting & 24{,}576 \\
Generation limit per turn & 16{,}384 \\
Checkpoint / validation interval & every 2 / every 10 updates \\
Maximum generation staleness & 1 update \\
Inference engines & 4, tensor-parallel size 1, bfloat16 \\
Random seed & 42 \\
\bottomrule
\end{tabularx}
\end{center}
\end{table}

\begin{table}[htbp]
\caption{Per-variant training configuration. The schedule for \GRPOPRMDecayed{} takes
$\lambda$ from 0.5 to zero with a cubic curve over updates 20 to 30 and holds it there
through update 50. \OPSD{} optimizes
the top-$K$ distillation objective of Equation~\ref{eq:opsd-kl} against a
frozen copy of the initial policy, retaining the teacher's top 64 tokens, and
uses no outcome or process reward. Our \OPSD{} runs sample one rollout per task
rather than eight, so their configured trajectory count per update is one
eighth of that of the reward-based variants.}
\label{tab:per-arm-config}
\setlength{\tabcolsep}{5pt}
\begin{center}
\begin{tabularx}{\linewidth}{@{}p{120pt}*{4}{>{\centering\arraybackslash}X}@{}}
\toprule
\multicolumn{5}{@{}l}{\textbf{\baseModel{}}} \\
\midrule
Setting & \shortstack{\OutcomeGRPO[\\]} & \shortstack{\GRPOPRMFixed[\\]} & \shortstack{\GRPOPRMDecayed[\\]} & \OPSD{} \\
\midrule
Rollouts per task & 8 & 8 & 8 & 1 \\
\ProcessRewardWeight{} $\lambda$ & 0 & 0.5 & $0.5\!\to\!0$ & 0 \\
Advantage estimator & \GRPO{} & \GRPO{} & \GRPO{} & --- \\
Policy loss & \multicolumn{3}{c}{rollout importance sampling} & top-$K$ KL \\
Zero-variance filter & yes & yes & yes & no \\
Context limit & \multicolumn{4}{c@{}}{131{,}072} \\
Policy GPUs & 4 & 4 & 4 & 4 \\
\bottomrule
\end{tabularx}
\end{center}
\end{table}

\paragraph{Policy optimization and advantage normalization.}
The reward-based variants use token-level importance ratios between the current
policy and the rollout policy. Ratios outside $(0.8,\,1.28)$ receive zero weight;
the retained ratios are treated as constants during backpropagation.
The policy loss uses token-mean reduction over valid assistant tokens.
For Equation~\ref{eq:prm-advantage}, only valid process scores from turns with
active loss tokens enter the group statistics, with $\epsilon=10^{-6}$.
The process advantage is zero for turns with missing or invalid scores, groups
with fewer than two valid scores, or groups with sample standard deviation at
most $\epsilon$; the outcome term is retained.

\paragraph{Process-weight schedule.}
The fixed variant uses $\lambda=0.5$ at every update. For the decayed variant,
the coefficient is $0.5$ through update 20, then
$0.5[1-(u-20)/10]^3$ for $20<u<30$, and zero from update 30 onward.
Here $u$ is the consumer optimizer's global update index, rather than the
index at which an asynchronous rollout began. For example, updates 21 and 25
use coefficients 0.3645 and 0.0625. Setting the coefficient to zero removes
the process term from the policy advantage; it does not change the outcome
reward or the final evaluation grader.

\paragraph{Token weighting and comparison scope.}
The token-mean loss divides the summed contributions of valid assistant tokens
by the total number of such tokens in the optimization batch. Thus each valid
token has the same reduction weight; turns and trajectories are not individually
renormalized to have equal weight. Prompt tokens, tool observations, padding,
and positions with an inactive loss mask do not contribute to this loss.
The configured batch targets eight trajectories for each of 16 tasks in
the reward-based variants and one trajectory for each of 16 tasks in each
\OPSD{} variant. Matching 50 updates therefore does not match the number of sampled
trajectories or generated tokens across these objectives. The comparisons in
this paper are comparisons of the specified training recipes.

\FloatBarrier
\subsection{Process-Reward Judgments}
\label{app:process-judge}

\paragraph{Context and timing.}
The process judge reviews consecutive windows of ten completed assistant turns;
the final window may be shorter. It receives the original public task,
public assistant actions and tool observations, registered images, and typed
facts derived from the available records. Separate reasoning/thinking fields
are excluded from this packet. Target turns and a bounded selection of earlier
turns retain detailed evidence; omitted earlier material is explicitly marked
as incomplete history. The image packet prioritizes the current window and
uses up to 16 original images, reserving at most four for earlier context.
These are judge-input limits, not limits on the policy's observations.

Each window receives three independent judgments from Gemini 3.8 Flash.
For a particular turn, valid scores are averaged; an abstention is excluded
from this average rather than treated as a neutral vote. If no valid score
remains, that turn receives no process contribution. The judge is asked to cite
only evidence available at or before the turn being assessed. Since several
target turns appear together in a retrospective window, this citation rule
does not establish that the model's judgment is causally independent of later
turns in that window.

\begin{table}[htbp]
\caption{Process-judgment decision rules. These summarize the implemented rubric;
they specify intended supervision rather than guarantee the correctness of every judgment.}
\label{tab:process-judge-rubric}
\begin{center}
\begin{tabularx}{\linewidth}{@{}p{43pt}X@{}}
\toprule
Score & Evidence required \\
\midrule
$+1$ & New task-relevant information or inference supported by observations;
a verified correction; or a substantive edit that demonstrably advances a task requirement. \\
$0$ & Routine file operations, an unchanged check, or an honest status update
without verified new progress. A successful command or valid JSON alone does not earn credit. \\
$-1$ & An observed error, a demonstrated task violation, an invalid claim of verification,
or a supported persistent repetition of already-resolved work. \\
Abstain & Material evidence is missing or inadequate for the proposed judgment.
Unavailable media or an infrastructure failure alone does not establish an agent error. \\
\bottomrule
\end{tabularx}
\end{center}
\end{table}

\paragraph{Evidence validation.}
Decisions identify the turn, event type, evidence scope, score, and supporting
record IDs. Validators check the allowed fields, event-to-score mapping,
record identities and time boundaries. Visual claims require attached pixels;
text metadata is not treated as direct auditory evidence. In the archived
process-judge adapter, authenticated audio attachments are unavailable, so
claims requiring direct listening must abstain unless the available evidence
supports a narrower metadata or reasoning judgment. Routine readback of an
unchanged JSON file is neutral. The repetition rule is deliberately narrow:
it requires the same literal shell command and matching observed text over
three consecutive turns, together with evidence that the work is redundant.
Such validation checks provenance and protocol compliance; it does not prove
semantic correctness or eliminate reward hacking.

\FloatBarrier
\subsection{Evaluation and Aggregation}
\label{app:evaluation-protocol}

\paragraph{Evaluation configuration.}
Table~\ref{tab:eval-config} lists the evaluation settings. The outcome grader
and the process-reward judge are different models, so process rewards are not
produced by the model that assigns the reported outcome scores.

\begin{table}[htbp]
\caption{Evaluation configuration. The outcome grader applies the rubric
formulation of Section~\ref{sec:trailer-problem-setup}; the process-reward judge is
used for training the \GRPOPRM{} variants and for the turn-quality analysis.}
\label{tab:eval-config}
\begin{center}
\begin{tabularx}{\linewidth}{@{}l>{\raggedright\arraybackslash}X@{}}
\toprule
Setting & Value \\
\midrule
Validation tasks & 81, from six held-out films \\
Evaluations per task & 4 independent repetitions \\
Evaluated checkpoint & after 50 updates, unless stated otherwise \\
Sampling & temperature 1.0, top-$p$ 1.0, top-$K$ $-1$ \\
Thinking & enabled \\
Turn budget & 100 \\
Max generated tokens per turn & 16{,}384 \\
Context limit & 131{,}072 \\
Outcome grader & Gemini 3.1 Pro Preview, temperature 0, seed 42 \\
Process-reward judge & Gemini 3.8 Flash, temperature 1 \\
Process judgments per turn & $M=3$, averaged over the valid ones \\
Process score per judgment & $-1$, $0$, or $+1$; abstentions are dropped \\
\bottomrule
\end{tabularx}
\end{center}
\end{table}

\paragraph{Score aggregation.}
For each method and task, avg@4 averages the four outcome scores, rather than
selecting the best attempt. Task-type scores then average equally over tasks of
that type. Overall averages equally over all 81 tasks: Content Selection has
14 tasks, Narrative Sequencing 15, Rough Cut 26, and Fine Cut 26. Overall is
therefore a task-count-weighted average of the four type scores, not their
unweighted mean. Relative gains over the base model use the
unrounded aggregated scores as $(R-R_{\mathrm{base}})/R_{\mathrm{base}}$;
the three-decimal scores displayed in the table may not reproduce these
percentages exactly. Absolute score differences are likewise computed before
rounding.

\paragraph{Interaction length.}
Table~\ref{tab:interaction-length} reports the mean number of turns by task type.
A turn is one policy request, counted from the rollout's recorded request IDs.
Each method includes all 324 rollouts from four independent repetitions of the
81 validation tasks, including rollouts whose outcome grades required recovery.

\begin{table}[htbp]
\caption{Mean turns by task type over four independent repetitions of all 81
validation tasks (324 rollouts per method). Overall averages over all tasks,
weighted by the number of tasks of each type.}
\label{tab:interaction-length}
\setlength{\tabcolsep}{4pt}
\begin{center}
\begin{tabularx}{\linewidth}{@{}l*{5}{>{\centering\arraybackslash}X}@{}}
\toprule
Method & CS & NS & RC & FC & Overall \\
\midrule
\baseModel{} & 22.3 & 15.3 & 21.8 & 87.2 & 41.7 \\
\OutcomeGRPO{} & 34.8 & 32.3 & 37.7 & 65.4 & 45.1 \\
\GRPOPRMFixed{} & 70.4 & 84.6 & 85.1 & 84.9 & 82.4 \\
\GRPOPRMDecayed{} & 31.6 & 52.5 & 54.3 & 75.7 & 56.9 \\
\OPSDReference{} & 23.0 & 13.5 & 22.6 & 85.2 & 41.1 \\
\OPSDSuggestions{} & 11.8 & 9.0 & 12.2 & 13.0 & 11.8 \\
\bottomrule
\end{tabularx}
\end{center}
\end{table}

\FloatBarrier
\subsection{Behavioral Measurements}
\label{app:behavior-measurement}

The behavioral analysis uses the four recorded validation repetitions of each
\baseModel{} variant. Each variant contributes 324 task--repetition pairs. We
count a turn from its policy-request identifier, so a request that is interrupted
before a completed generation still contributes to interaction length. Tool calls
count requested invocations, whereas frames count image attachments returned in
completed tool results; these need not increase together. A contact sheet is one
image attachment even when it contains multiple sampled frames. Neither measure
counts distinct source-film frames. Blind rollouts are those with no returned
image attachment. Truncation indicates a recorded budget termination, rather
than an assessment of whether the final artifact is correct.

\paragraph{Reasoning length.}
Reasoning length is measured in \emph{characters}, not tokenizer tokens. For each
completed generation event, we parse the action text as JSON and read its
\texttt{reasoning} field. A missing or non-string field contributes zero characters.
We divide the total number of reasoning characters by the number of completed
generation events, pooling all tasks and repetitions within a method. This is a
turn-weighted average, not the average of per-rollout ratios. It is distinct from
the number of policy requests used for interaction length, because an interrupted
request may not emit a completed generation event.

\paragraph{Reflection markers.}
We apply the case-insensitive expression below to the same reasoning field. A
completed turn counts once if any alternative matches, even if several markers
occur. The reflection rate is the number of matching turns divided by the number
of completed generation events, again pooling tasks and repetitions. Missing
reasoning is a non-match. The expression uses Python regular-expression syntax;
\texttt{\textbackslash b} is a word boundary and \texttt{\textbackslash s} is whitespace.

\begingroup\normalsize
\begin{verbatim}
re.compile(
    r"\bwait[,.\s]"
    r"|\breconsider\b"
    r"|\blet me (re-?check|re-?examine|re-?visit|re-?think|"
    r"re-?do|re-?count|re-?calculate|re-?verify|"
    r"retry|try again)\b"
    r"|\b(was|were|is|are)\s+(wrong|incorrect|invalid)\b"
    r"|\bmy (mistake|error)\b"
    r"|\bshould have\b"
    r"|\b(did|does)\s?n[o']?t work\b"
    r"|\btry again\b"
    r"|\bon second thought\b"
    r"|\bthat'?s not (right|correct)\b"
    r"|\b(previous|earlier|last) (attempt|approach|try|call)s?\b"
    r"|\bI (over|mis)[a-z]+", re.I)
\end{verbatim}
\endgroup

This is a lexical indicator of explicit reconsideration. It does not establish
that the agent found an error, changed its next action, or improved the edit.
For example, an isolated ``wait'' or a mention of a previous attempt can match
without an effective correction. Longer reasoning also provides more
opportunities for a match. Consequently, the behavior columns describe observed
policy activity and should be interpreted together with task outcome and
completion, rather than as independent measures of reasoning quality.

\FloatBarrier
\subsection{OPSD Teacher Inputs and Diagnostics}
\label{app:opsd-implementation}

\paragraph{Teacher and student.}
The KL teacher is a frozen copy of the initial \baseModel{}, initialized from
the same checkpoint as the student. It is not updated by an exponential moving
average or replaced by the trained student. The student samples its own actions
from the public task context. Teacher-side privileged information changes the
distribution evaluated at those same student action tokens; it is not inserted
into student rollout or evaluation inputs, and it does not replace student
actions with supervised action labels.

\paragraph{Reference-answer conditioning.}
The private answer card contains the selected clip IDs for Content Selection;
the reference order for Narrative Sequencing; and clip-relative reference
in/out intervals and target duration for Rough Cut. For Fine Cut it contains
reference join decisions, dialogue identities and onset times, and a music
source offset. The Fine Cut card is explicitly a partial editing plan, not a
ready-to-submit executable edit: unprovided trim endpoints, gains, mixing
envelopes and render commands are not invented as reference targets.
Rough Cut and Fine Cut reference choices are marked non-unique. The card is
inserted only into the teacher's existing system message while preserving the
student's action-token and image suffix. Cards are bound to training task
identities and the corresponding stored answer and manifest; validation
tasks are excluded from this private catalog.

\paragraph{Turn-suggestion conditioning.}
Gemini 3.8 Flash receives the original public context immediately before turn
$t$, including prior tool observations and admitted images, and proposes one
next action. It neither executes that action nor observes the student's current
action or future outcome. An action may contain multiple tool calls. The
proposal is appended as a teacher-only note after the original turn context,
followed by the original student action tokens. A separate teacher prefix is
constructed for each turn; the note from a previous or future turn is not
carried into it. Gemini is the suggestion generator, while the frozen
\baseModel{} remains the model that supplies KL targets. Suggestions are
checked for format, available tool names, argument schemas and a maximum of
2,048 teacher-hint tokens. These checks do not establish that a suggestion is
semantically correct or executable in the student's state.

\paragraph{Teacher context and objective.}
Both variants reserve up to 2,048 additional teacher-only tokens, giving a
teacher context limit of 133,120 tokens while the student's limit remains
131,072. The implemented forward-KL objective uses neither an outcome-reward
term nor a process-reward term, and does not apply sampled-action importance
weighting. Outcomes are still recorded for evaluation and diagnostics.

\paragraph{Top-$K$+tail KL in \OPSD{}.}
\label{app:opsd-topk-kl}
We give the implemented form of Equation~\ref{eq:opsd-kl} here. Let
$p_{t,k}$ and $q_{t,k}$ be its full-vocabulary student and teacher
distributions at token position $k$ of turn $t$. We use uppercase $K=64$
for the support size, to distinguish it from the token index $k$. At each
position, the teacher selects
\begin{equation}
  S_{t,k}=\operatorname{TopK}_{v\in\mathcal V} q_{t,k}(v),
  \qquad |S_{t,k}|=K.
  \label{eq:opsd-topk-support}
\end{equation}
Both models are evaluated on this same teacher-selected set; the student
does not select a separate top-$K$ set. The selection is made independently
at each prediction position on the student's sampled prefix.

All retained probabilities keep their respective full-vocabulary softmax
normalizers. We aggregate the remaining vocabulary into one tail class
$\bot$, with
\begin{equation}
  q_{t,k}^{\bot}=\sum_{v\notin S_{t,k}}q_{t,k}(v),
  \qquad
  p_{t,k}^{\bot}=\sum_{v\notin S_{t,k}}p_{t,k}(v).
  \label{eq:opsd-tail-mass}
\end{equation}
Thus $\widetilde q_{t,k}=((q_{t,k}(v))_{v\in S_{t,k}},q_{t,k}^{\bot})$
and $\widetilde p_{t,k}=((p_{t,k}(v))_{v\in S_{t,k}},p_{t,k}^{\bot})$
are normalized distributions over the same $K+1=65$ classes. We do not
renormalize the retained 64 entries to sum to one or discard the tail mass.
The per-token forward KL used in both \OPSD{} variants is
\begin{equation}
\begin{aligned}
  \ell_{t,k}^{(K)}
  &=D_{\mathrm{KL}}(\widetilde q_{t,k}\|\widetilde p_{t,k})\\
  &=\sum_{v\in S_{t,k}}q_{t,k}(v)
      \log\frac{q_{t,k}(v)}{p_{t,k}(v)}
    +q_{t,k}^{\bot}\log\frac{q_{t,k}^{\bot}}{p_{t,k}^{\bot}}.
\end{aligned}
\label{eq:opsd-topk-tail-kl}
\end{equation}
The direction is teacher-to-student, $D_{\mathrm{KL}}(q\|p)$.
The student-sampled token need not belong to $S_{t,k}$: this is a
distribution-level target at every valid assistant prediction position,
not a loss evaluated only at the sampled token.

For an optimization batch $\mathcal B$, let $m_{i,t,k}\in\{0,1\}$ denote
the valid assistant-token loss mask for trajectory $i$. The implemented
token-mean reduction is
\begin{equation}
  \mathcal L_{K}(\theta)=
  \frac{\sum_{i\in\mathcal B}\sum_{t,k}
    m_{i,t,k}\,\ell_{i,t,k}^{(K)}}
       {\sum_{i\in\mathcal B}\sum_{t,k}m_{i,t,k}}.
  \label{eq:opsd-topk-token-mean}
\end{equation}
Prompt tokens, tool observations and padding are excluded. Turns and
trajectories are not separately normalized to have equal weight. Teacher
targets, selected vocabulary IDs and sampled prefixes are held fixed
within an update; gradients pass through the student probabilities,
including their full-vocabulary normalizer and aggregate tail mass.

\paragraph{Approximation and numerical implementation.}
The top-$K$+tail construction preserves each selected probability and the
total probability outside the selected set. It does not preserve how
probability is distributed among individual tail tokens. Its KL is
therefore a coarsened objective, rather than the exact full-vocabulary KL;
it is no greater than the full KL and becomes equal when the conditional
teacher and student distributions within the tail agree.

This construction reduces each stored teacher target to 64 vocabulary IDs
and 65 log-probabilities per active position. It does not avoid the
full-vocabulary projection needed for normalization. For numerical
stability, the implementation computes each tail log-probability as the
log-sum-exp of the complementary logits minus the full-vocabulary
log-sum-exp, rather than subtracting a near-one retained probability
from one. Logits are projected in chunks of 128 active positions, and the
backward pass recomputes these projections instead of retaining a
trajectory-by-vocabulary tensor. Log-probability arithmetic uses FP32 in
the reported runs. We use the same $K=64$ in both variants and do not
claim that this value is optimal or that tail aggregation preserves all
full-vocabulary supervision.

\paragraph{Interpreting teacher--student disagreement.}
Changing privileged information changes the teacher targets on student-visited
states. A larger KL identifies stronger distributional disagreement, but does
not by itself establish that the teacher proposes a better action. Likewise,
a larger parameter-gradient norm is an observed optimization difference, not
proof that a particular probability ratio causes instability. The top-$K$+tail
forward-KL implementation differentiates through normalized student logits;
for a retained vocabulary entry, its logit derivative is the difference
between student and teacher probabilities. We therefore interpret the loss,
gradient and outcome measurements together, without treating the derivative
with respect to an unconstrained probability as a parameter-gradient mechanism.

\paragraph{Gradient magnitude by \privilegedInformation{}.}
Table~\ref{tab:gradient-norm} reports the two \OPSD{} mean gradient norms used in
Section~\ref{sec:analysis-opsd}, alongside additional diagnostics from the
50 retained optimizer updates. Gradient norms are measured before global clipping. The two \OPSD{} variants and \outcomeGRPO{} share the
learning rate and the clipping threshold. Their objectives, targets, sampled
trajectories and evolving policies can all affect the observed norms; this
comparison does not isolate a single cause.

\begin{table}[htbp]
\caption{Mean teacher--student KL per token and gradient-norm summaries.
The two \OPSD{} mean gradient norms are those reported in
Section~\ref{sec:analysis-opsd}. Mean KL, maximum norms, clipping counts and
the \outcomeGRPO{} mean norm are computed over the 50 retained optimizer
updates, excluding superseded
restart attempts. Gradient norms are measured before global clipping. Rows
are ordered by the mean norm. The clipping threshold is $1.0$, and the
Clipped column counts retained updates whose norm exceeded it. \OutcomeGRPO{}
trains on no distillation target and therefore has no teacher--student KL, and its
row pools the 50 updates of that run across its resumed segments.}
\label{tab:gradient-norm}
\setlength{\tabcolsep}{5pt}
\begin{center}
\begin{tabular}{@{}lcccc@{}}
\toprule
\PrivilegedInformation{} & Mean KL & Mean norm & Max norm & Clipped \\
\midrule
\ReferenceAnswers{} & 0.0019 & 0.084 & 0.249 & 0 \\
\textit{\OutcomeGRPO{}} & \textit{---} & \textit{0.240} & \textit{0.296} & \textit{0} \\
\TurnSuggestions{} & \textbf{0.0279} & \textbf{1.177} & \textbf{3.981} & \textbf{22} \\
\bottomrule
\end{tabular}
\end{center}
\end{table}

\paragraph{How the divergence is spread over a trajectory.}
Table~\ref{tab:kl-concentration} uses the 800 trajectories stored in the
pre-update signal records (16 per update for 50 updates) of each reported
\OPSD{} run. Turns are ranked by their summed token KL, and the table
reports the share of total KL held by the top 1\% and 5\% of turns, and
the token share held by the top 5\% (rounded down to an integer number of turns). For the last
column, turn $j$ in a trajectory of length $L$ has relative position
$j/\max(1,L-1)$ with zero-based $j$. We average each turn's per-token KL
within the earliest ($<0.2$) and latest ($\geq0.8$) fifths and take their
ratio. This diagnostic weights turns equally within each positional bin;
it does not change the token-mean training objective.

With \referenceAnswers{}, a median recorded trajectory contains $23$ turns
and $13{,}680$ generated tokens. Under token-mean reduction, concentrating
disagreement in relatively few turns may dilute their contribution when
averaged with the remaining tokens. These statistics do not establish
whether the divergent turns provide useful guidance or whether this
concentration causes the limited performance gains.

\begin{table}[htbp]
\caption{Concentration of the teacher--student KL across the turns of a run, over
the 800 recorded training trajectories per run. Ranking uses total KL
per turn; the positional ratio uses the mean of per-turn, per-token KL values.}
\label{tab:kl-concentration}
\setlength{\tabcolsep}{5pt}
\begin{center}
\begin{tabular}{@{}lcccc@{}}
\toprule
& \multicolumn{2}{c}{Share of total KL held by} & & First fifth \\
\cmidrule(lr){2-3}
\PrivilegedInformation{} & top $1\%$ turns & top $5\%$ turns & Tokens in top $5\%$ & over last \\
\midrule
\ReferenceAnswers{} & 19.8\% & 47.4\% & 17.9\% & 3.91 \\
\TurnSuggestions{} & 5.6\% & 19.6\% & 14.2\% & 0.72 \\
\bottomrule
\end{tabular}
\end{center}
\end{table}

\paragraph{Does the target track task success.}
For every rollout of every update we take its mean KL per token against the
teacher and the outcome reward it earns, which gives $800$ paired values per run
over the 50 updates. Table~\ref{tab:kl-outcome} reports the Pearson correlation
between the two. This is a descriptive association between distributional disagreement and
realized outcome. The observed correlation is modestly negative for
\referenceAnswers{} and near zero for \turnSuggestions{}. Correlation alone
neither establishes teacher correctness nor tests whether an individual
teacher-guided update improves the student.

\begin{table}[htbp]
\caption{Correlation between the per-token KL of a rollout and the outcome reward
it earns, over the $800$ rollouts of each run. The reported $t$ is the nominal Pearson-correlation statistic. These pooled
rollouts share training updates and task identities; the statistic does not
account for this dependence and should not be interpreted as a causal test.}
\label{tab:kl-outcome}
\setlength{\tabcolsep}{6pt}
\begin{center}
\begin{tabular}{@{}lccc@{}}
\toprule
\PrivilegedInformation{} & $r$ & $r^2$ & $t$ \\
\midrule
\ReferenceAnswers{} & $-0.165$ & 0.027 & $-4.71$ \\
\TurnSuggestions{} & $-0.007$ & 0.000 & $-0.19$ \\
\bottomrule
\end{tabular}
\end{center}
\end{table}

\FloatBarrier
\subsection{Outcome Metrics and Rubrics}
\label{app:outcome-scoring}

\paragraph{Outcome scoring.}
The channel weights $(w_{\mathrm{det}}, w_{\mathrm{ref}}, w_{\mathrm{blind}})$ of
Equation~\ref{eq:outcome-reward} are $(1,0,0)$ for Content Selection and Narrative
Sequencing, $(0.4,0.4,0.2)$ for Rough Cut, and $(0,0.4,0.6)$ for Fine Cut.

\paragraph{Deterministic metrics.}
Content Selection uses set F1, $2|S\cap S^\star|/(|S|+|S^\star|)$, on the
submitted and reference clip IDs. Narrative Sequencing maps the submitted
permutation to reference ranks and multiplies three terms: one minus normalized
absolute displacement, the length of the longest increasing subsequence divided
by the number of clips, and the fraction of consecutive submitted pairs that
are consecutive in the reference~\citep{cao2026agenticvbench}. Absolute
displacement is normalized by $\lfloor n^2/2\rfloor$ for $n$ clips. The
Heretic c5 verifier used in our experiments takes the best score over its four accepted orders.
Kendall's $\tau$ is logged as a diagnostic but is not the outcome score.

For Rough Cut, a scorable shot is a hit only if both its in-point error and
its retained-duration error fall within the corresponding reference
tolerances. The deterministic score is the fraction of scorable shots that
are hits. In-point tolerances are shot-specific, derived from the alignment's
self-similarity plateau; retained-duration tolerance comes from the task key.
Unmatched alternate takes remain in the delivered sequence but are excluded
from this reference-based denominator. Invalid ranges are not exempted as
unscorable shots. Fine Cut has no deterministic similarity channel.

The two judge channels are rubric-based, and Table~\ref{tab:rubrics} lists their items. A
separate multimodal judgment grades each subjective item from 0 to 4 using the
submitted continuous video and the relevant evidence. Rough Cut additionally
has a mechanically assigned duration band in its reference channel. The item
scores combine into a channel score by
\begin{equation}
    R_{\mathrm{channel}}
      = \frac{\sum_i s_i w_i}{4\sum_i w_i}\,(0.75)^{z},
    \label{eq:rubric-aggregation}
\end{equation}
where $z$ counts the items that scored 0. The factor $(0.75)^{z}$ is a soft
penalty. The final evaluation uses direct rubric judgments with submission
validation enabled; no additional joint consistency-review call modifies the
channel scores. A complete Rough Cut grade comprises six subjective rubric
judgments plus mechanical metrics; a Fine Cut grade comprises seven subjective
rubric judgments. These counts describe distinct items, not provider requests
after transport or output-format retries. Missing required grades are recorded
as grading failures rather than silently renormalizing the remaining channels.

\begin{table}[htbp]
\caption{Rubric items and weights inside each judge channel. The blind channel asks
whether the edit works on its own terms, and the reference channel asks how far it departs
from the released trailer.}
\label{tab:rubrics}
\begin{center}
\begin{tabularx}{\linewidth}{@{}ll>{\raggedright\arraybackslash}X@{}}
\toprule
Type & Channel & Items with weights \\
\midrule
RC & Blind & cut placement 0.35, beat coverage 0.40, join continuity 0.25 \\
RC & Reference & picture entries 0.24, picture exits 0.24, pacing 0.32, mechanical duration 0.20 \\
FC & Blind & dialogue 0.32, music 0.20, continuity 0.20, transition 0.28 \\
FC & Reference & dialogue 0.40, continuity 0.35, music 0.25 \\
\bottomrule
\end{tabularx}
\end{center}
\end{table}


\paragraph{What each rubric evaluates.}
Tables~\ref{tab:rc-rubric-details} and~\ref{tab:fc-rubric-details} specify the
individual criteria. The task prompt establishes the required visual beat and
audiovisual functions. The public manifest identifies the available assets and
their time coordinates. Reference-informed judgments additionally receive
reference evidence, but a reference choice does not by itself make that choice
mandatory. A different edit can receive full credit when it satisfies the task.
Judges must ground their assessment in the submitted continuous video and audio,
using clip or join IDs and timestamps to locate the evidence. Contact sheets,
source context, ASR, and signal measurements support inspection; they do not
replace watching and listening to the submission. A missing measurement means
unknown, and an empty defect list is not sufficient evidence of quality.

\begin{table}[htbp]
\setlength{\tabcolsep}{4pt}
\renewcommand{\arraystretch}{1.15}
\caption{RC rubric details. B and R denote judgment without the reference
and reference-informed judgment. Weights are within the respective channel;
the R weights include the mechanical duration item. Source context can contain
discarded footage and artificial reel joins, neither of which is submitted
content.}
\label{tab:rc-rubric-details}
\small
\begin{center}
\begin{tabularx}{\linewidth}{@{}>{\raggedright\arraybackslash}p{0.21\linewidth}*{2}{>{\raggedright\arraybackslash}X}@{}}
\toprule
Item and weight & Evidence and task requirement & Failure criteria and scope \\
\midrule
B: cut placement (0.35) &
Submitted picture entries and exits, inspected with the mapped source
cut-context reel; whether the selected windows make the intended actions
legible. &
Visible premature cuts, truncated actions, missed moments, or dead holds.
Equal shot lengths, starts at zero, or static similarity alone do not show a
defect. Finished sound is not required. \\
\addlinespace
B: visual coverage (0.40) &
The full assembled sequence against the prompt's explicit visual actions,
relationships, and requested beat. &
Required content is absent, unclear, or too brief to understand. An action
visible only in discarded source context does not count as delivered; ASR does
not substitute for visual coverage. \\
\addlinespace
B: join continuity (0.25) &
Actual consecutive submitted shots in motion; spatial and action continuity,
rhythm, and the style requested by the task. &
Distracting discontinuities or confusing joins that disrupt the intended
sequence. Static likeness is not proof of a jump cut, and deliberate montage
can be appropriate. \\
\addlinespace
R: picture entries (0.24) &
Submitted entries, source context, and available reference entry evidence;
whether useful openings and essential picture moments remain legible. &
Entries that miss the useful opening or linger before it. A numeric departure
from the reference is not itself a defect; effective alternative entries can
receive full credit. \\
\addlinespace
R: picture exits (0.24) &
Submitted exits and available reference exit evidence; whether actions have
enough time to read without unnecessary tails. &
Visibly clipped actions or prolonged dead tails. Credit depends on the actual
loss or preservation of the shot's purpose, not exact reference matching. \\
\addlinespace
R: pacing (0.32) &
Continuous playback, the requested beat, and available reference pacing
evidence; the distribution of time and emphasis across shots. &
Concrete rushing, stalling, or misplaced emphasis. Total duration is scored
separately; uniform lengths or a different duration-variation statistic are
not faults by themselves. \\
\addlinespace
R: total duration (0.20; mechanical) &
The sum of submitted kept intervals versus the answer key's
\texttt{target\_duration\_s}, with an absolute tolerance floor from
\texttt{dur\_tol\_s}. &
Bands 4, 3, 2, and 1 allow absolute errors up to 3\%, 8\%, 18\%, and 35\%
of the target, respectively, each with the absolute floor applied; larger
errors receive 0. This does not replace the submission-validity gate. \\
\bottomrule
\end{tabularx}
\end{center}
\end{table}

For the mechanical RC duration item, each percentage threshold is replaced by
the larger of that percentage of the target and the answer key's
\texttt{dur\_tol\_s}; the implementation uses 0.1668 seconds if that field is
absent or zero. The band is computed arithmetically without an LLM call.
This item measures total duration, whereas the separate deterministic RC score
tests both in-point and kept-duration tolerances shot by shot. These two uses
of timing should not be conflated with the validity checks applied before
judging.

\begin{table}[htbp]
\setlength{\tabcolsep}{4pt}
\renewcommand{\arraystretch}{1.08}
\caption{FC rubric details. B and R denote judgment without the reference
and reference-informed judgment. Weights are within each channel. All items
evaluate the delivered audiovisual result: a declared JSON operation or a
possible source-media alignment is not evidence that it was realized.}
\label{tab:fc-rubric-details}
\small
\centering
\begin{tabularx}{\linewidth}{@{}>{\raggedright\arraybackslash}p{0.21\linewidth}*{2}{>{\raggedright\arraybackslash}X}@{}}
\toprule
Item and weight & Evidence and task requirement & Failure criteria and scope \\
\midrule
B: dialogue (0.32) &
Listen to the submitted mix while watching the picture; assess intelligibility,
placement, and narrative relevance against the prompt's explicit requirements. &
Required speech is masked, missing, irrelevant, or poorly placed. Missing ASR
words alone do not prove inaudibility; lip-sync claims require visible mouth
and audible speech evidence. Appropriate absence is allowed. \\
\addlinespace
B: music (0.20) &
The audible cue's entry, exit, phrasing, balance, and relationship to the
picture and dialogue, where the brief requires those functions. &
Music fights the picture, masks speech, or fails a required musical function.
A declared cue offset is not delivered alignment. A steady bed or silence can
be correct; there is no requirement to hit every cut. \\
\addlinespace
B: continuity choices (0.20) &
Actual picture--sound continuity across joins; whether hard cuts, sustained
beds, dialogue overlaps, or other bridges serve the requested stretch. &
Observed continuity weaknesses disrupt the flow. There is no bridge quota:
one bed may span many joins, and a deliberate hard cut may be appropriate.
Unconfirmed crossings are not automatically missing bridges. \\
\addlinespace
B: transition execution (0.28) &
The continuous render at actual transition locations, with measured
discontinuities used only to locate possible problems. &
Audible chops, distracting level jumps, unintended black, inappropriate fades,
or disruptive seams. Source black, title cards, and required pauses are not
errors; signal discontinuities alone do not establish defects. \\
\addlinespace
R: dialogue (0.40) &
The same delivered-dialogue evidence, informed by the reference's dialogue
choices and the task's intended narrative function. &
An actual loss of required meaning, intelligibility, or effective placement.
A different line selection is not penalized solely for differing from the
reference or leaving a source opportunity unused. \\
\addlinespace
R: continuity (0.35) &
Delivered continuity choices and their execution, considered with the
available reference evidence for the intended beat. &
Disruptive joins or transition defects that weaken the actual sequence.
Reference bridge annotations describe one solution; they do not impose a
transition-count target or require identical bridge types. \\
\addlinespace
R: music (0.25) &
The audible mix and its relationship to picture and dialogue, informed by the
available reference musical treatment and the prompt. &
An actual phrasing, balance, entry/exit, or task-relevance problem. Numeric
distance from the reference cue offset alone is insufficient; an effective
alternative cue placement can receive full credit. \\
\bottomrule
\end{tabularx}
\end{table}

The subjective rubrics distinguish full satisfaction from minor, repeated,
major, and fundamental failures using the criterion-specific conditions in
these tables. They do not introduce a common numeric threshold for dialogue
loudness, bridge counts, shot-length variation, or source-reference distance.
The highest band requires positive evidence that the stated criterion is met;
a lower band requires an observable, task-relevant defect. Reference media is
used only when actually available to the judge, and source annotations or
declared edit notes cannot establish that their contents occur in the submitted
render.

\paragraph{Rubric interpretation.}
The blind Rough Cut rubric assesses useful picture entry/exit points, coverage
of the requested visual beat, and continuity of the actual joins. Its
reference channel assesses entries, exits and pacing with the released edit
as a feasible comparison. Fine Cut evaluates delivered dialogue, music,
audiovisual continuity choices and transition execution; its reference
channel revisits dialogue, continuity and music. In all subjective rubrics,
band 4 denotes full task-appropriate compliance, band 3 a minor defect, band 2
several or a meaningful defect, band 1 substantial failure, and band 0 failure
of an essential requirement. The per-item criteria specify the relevant
visible or audible defect; this summary is not an additional scoring rule.
Appropriate silence, a hard cut, or an alternative edit is not penalized
merely for differing from the reference. Declarations, tool counts, source
transcripts and metadata alone do not prove a successful delivered result.

\paragraph{Rough Cut duration band.}
Let $T$ be the target duration and $E$ the absolute error in the summed retained
clip durations. The mechanical reference-channel item assigns bands 4, 3, 2,
and 1 when $E$ is at most 3\%, 8\%, 18\%, and 35\% of $T$, respectively;
each threshold has an absolute floor equal to the task's frame-based duration
tolerance. Larger errors receive band 0. This graded item is separate from
the common submission-validity gate, which allows total-duration error
$\max(1,\min(3,0.1T))$ seconds. A submission outside that gate receives zero
outcome reward before subjective quality can contribute.

\paragraph{Scoring exclusions.}
The protocol accepts the first valid grade rather than the best available one.
In the initial evaluation pass, judge refusals are not retried. Both rules avoid selecting for trajectories
that happen to score well on a later attempt. Refusals are rare but not
absent: a small number of Rough Cut trajectories trigger the judge's content
filter, which excludes that trajectory and the group it belongs to. Where this
leaves a repetition incomplete we average over the remaining complete
repetitions and report the count. For the six \baseModel{} variants reported in
Table~\ref{tab:main-results}, recovered grades fill the initially excluded evaluations,
giving complete coverage of all 324 task--repetition pairs per variant.
Behavioral metrics for \baseModel{} cover all four repetitions of the 81
validation tasks (Table~\ref{tab:interaction-behaviour}).

\FloatBarrier
\subsection{Additional Training Diagnostics}
\label{app:training-dynamics}

The following curves use optimizer-update metric records. Their logged generation
counts can differ from the fixed-size OPSD signal exports analyzed above;
interaction length uses the logged batch denominator. Averages over update
ranges give each recorded update equal weight.

\begin{figure}[htbp]
\centering
\definecolor{seriesproc}{HTML}{2A78D6}
\definecolor{seriesout}{HTML}{EB6834}
\definecolor{seriesdecay}{HTML}{007F83}
\definecolor{seriesdecayraw}{HTML}{C6E0E1}
\begin{tikzpicture}
\begin{axis}[
  width=0.95\linewidth, height=6cm,
  xlabel={Training update}, ylabel={Reward},
  xmin=0, xmax=51, ymin=-0.15, ymax=0.9,
  tick label style={font=\normalsize}, label style={font=\normalsize},
  legend style={font=\normalsize, at={(0.02,0.98)}, anchor=north west,
                draw=none, fill=none, row sep=0pt},
  grid=major, grid style={gray!18, line width=0.3pt},
  axis line style={gray!45, line width=0.4pt},
  tick style={gray!45},
]
\draw[gray!55, dashed, line width=0.6pt] (axis cs:30,-0.15) -- (axis cs:30,0.9);
\node[font=\normalsize, text=gray!75, anchor=west] at (axis cs:31,0.05)
  {reward hacking};
\addplot[forget plot, seriesproc, opacity=0.28, line width=0.4pt] coordinates {(1,0.0421) (2,-0.0828) (3,-0.0342) (4,-0.0828) (5,0.0315) (6,-0.0041) (7,0.0478) (8,-0.0001) (9,0.0637) (10,-0.0081) (11,0.0920) (12,0.1896) (13,0.2320) (14,0.2213) (15,0.3023) (16,0.1954) (17,0.2414) (18,0.3458) (19,0.3144) (20,0.4658) (21,0.2233) (22,0.3710) (23,0.2813) (24,0.4312) (25,0.3289) (26,0.5046) (27,0.4240) (28,0.5861) (29,0.5969) (31,0.5690) (32,0.4774) (33,0.5861) (34,0.5493) (35,0.5380) (36,0.6068) (37,0.5985) (38,0.6711) (39,0.6805) (40,0.7139) (41,0.8100) (42,0.7216) (43,0.7114) (44,0.7853) (45,0.7028) (46,0.7584) (47,0.7497) (48,0.7645) (49,0.7013) (50,0.7673)};
\addplot[forget plot, seriesout, opacity=0.28, line width=0.4pt] coordinates {(1,0.2224) (2,0.2293) (3,0.2502) (4,0.1868) (5,0.2346) (6,0.3654) (7,0.2134) (8,0.1815) (9,0.3507) (10,0.2167) (11,0.3361) (12,0.3045) (13,0.3028) (14,0.3484) (15,0.3659) (16,0.3264) (17,0.2823) (18,0.3405) (19,0.3210) (20,0.3438) (21,0.4144) (22,0.3849) (23,0.3393) (24,0.3306) (25,0.3398) (26,0.3584) (27,0.3349) (28,0.3631) (29,0.3712) (31,0.2994) (32,0.2467) (33,0.1950) (34,0.2262) (35,0.1988) (36,0.1714) (37,0.1218) (38,0.1658) (39,0.1566) (40,0.1509) (41,0.1114) (42,0.1207) (43,0.2280) (44,0.2282) (45,0.2334) (46,0.3181) (47,0.3452) (48,0.4440) (49,0.3145) (50,0.2292)};
\addplot[forget plot, seriesdecayraw, line width=0.4pt] coordinates {(21,0.39156016) (22,0.35958047) (23,0.35638984) (24,0.32080469) (25,0.40303438) (26,0.35069688) (27,0.37858281) (28,0.41621562) (29,0.46026719) (31,0.42126016) (32,0.42133382) (33,0.34433281) (35,0.47273516) (36,0.42538750) (37,0.41525000) (38,0.28562578) (39,0.25556641) (40,0.43718828) (41,0.32509922) (42,0.32126484) (43,0.41215139) (44,0.46705781) (45,0.27446016) (46,0.40363026) (47,0.42552734) (48,0.43988437) (49,0.41026912) (50,0.45894844)};
\addplot[seriesproc, line width=1.1pt] coordinates {(1,-0.0250) (2,-0.0394) (3,-0.0252) (4,-0.0345) (5,-0.0084) (6,-0.0015) (7,0.0278) (8,0.0198) (9,0.0391) (10,0.0674) (11,0.1138) (12,0.1453) (13,0.2074) (14,0.2281) (15,0.2385) (16,0.2612) (17,0.2799) (18,0.3126) (19,0.3181) (20,0.3441) (21,0.3312) (22,0.3545) (23,0.3271) (24,0.3834) (25,0.3940) (26,0.4550) (27,0.4881) (28,0.5361) (29,0.5307) (31,0.5631) (32,0.5557) (33,0.5439) (34,0.5515) (35,0.5757) (36,0.5927) (37,0.6190) (38,0.6542) (39,0.6948) (40,0.7194) (41,0.7275) (42,0.7484) (43,0.7462) (44,0.7359) (45,0.7415) (46,0.7521) (47,0.7353) (48,0.7482) (49,0.7457) (50,0.7443)};
\addlegendentry{Process reward (fixed $\lambda$)}
\addplot[seriesout, line width=1.1pt, dashed] coordinates {(1,0.2340) (2,0.2222) (3,0.2247) (4,0.2533) (5,0.2501) (6,0.2363) (7,0.2691) (8,0.2655) (9,0.2597) (10,0.2779) (11,0.3022) (12,0.3017) (13,0.3315) (14,0.3296) (15,0.3251) (16,0.3327) (17,0.3272) (18,0.3228) (19,0.3404) (20,0.3609) (21,0.3607) (22,0.3626) (23,0.3618) (24,0.3506) (25,0.3406) (26,0.3454) (27,0.3535) (28,0.3454) (29,0.3231) (31,0.2951) (32,0.2677) (33,0.2332) (34,0.2076) (35,0.1827) (36,0.1768) (37,0.1629) (38,0.1533) (39,0.1413) (40,0.1411) (41,0.1535) (42,0.1678) (43,0.1843) (44,0.2257) (45,0.2706) (46,0.3138) (47,0.3310) (48,0.3302) (49,0.3332) (50,0.3292)};
\addlegendentry{Outcome reward (fixed $\lambda$)}
\addplot[seriesdecay, line width=1.1pt] coordinates {(21,0.36917682) (22,0.35708379) (23,0.36627391) (24,0.35810125) (25,0.36190172) (26,0.37386687) (27,0.40175937) (28,0.40540453) (29,0.41953192) (31,0.41268192) (32,0.42398583) (33,0.41700989) (35,0.41580786) (36,0.38866625) (37,0.37091297) (38,0.36380359) (39,0.34374594) (40,0.32494891) (41,0.35025403) (42,0.39255231) (43,0.36000668) (44,0.37571289) (45,0.39656539) (46,0.40211199) (47,0.39075425) (48,0.42765191) (49,0.43365732) (50,0.43636731)};
\addlegendentry{Outcome reward (decayed $\lambda$)}
\end{axis}
\end{tikzpicture}
\caption{Training dynamics of \GRPOPRMFixed{} and \GRPOPRMDecayed{}.
Faint lines show recorded values; thick lines show centered five-observation
moving averages. With a fixed weight, the process reward keeps rising while
the outcome reward falls after update 30. The decayed-weight outcome curve
uses the same observations as Figure~\ref{fig:teaser}(c), starting at update 21;
lines connect available observations.}
\label{fig:training-dynamics}
\end{figure}
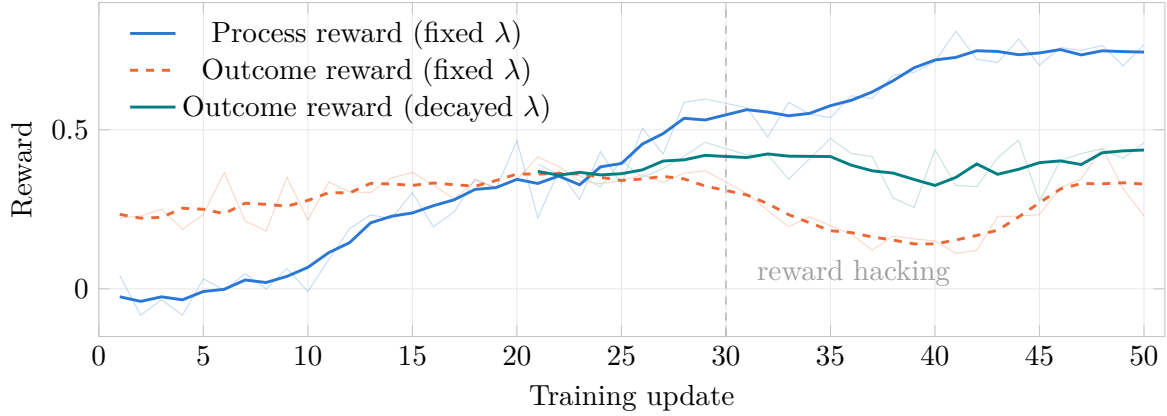

\paragraph{Process-reward dynamics.}
Figure~\ref{fig:training-dynamics} expands the process- and outcome-reward curves
shown in Figure~\ref{fig:teaser}(c).

\begin{figure}[tbp]
  \centering
  \includegraphics[width=\linewidth]{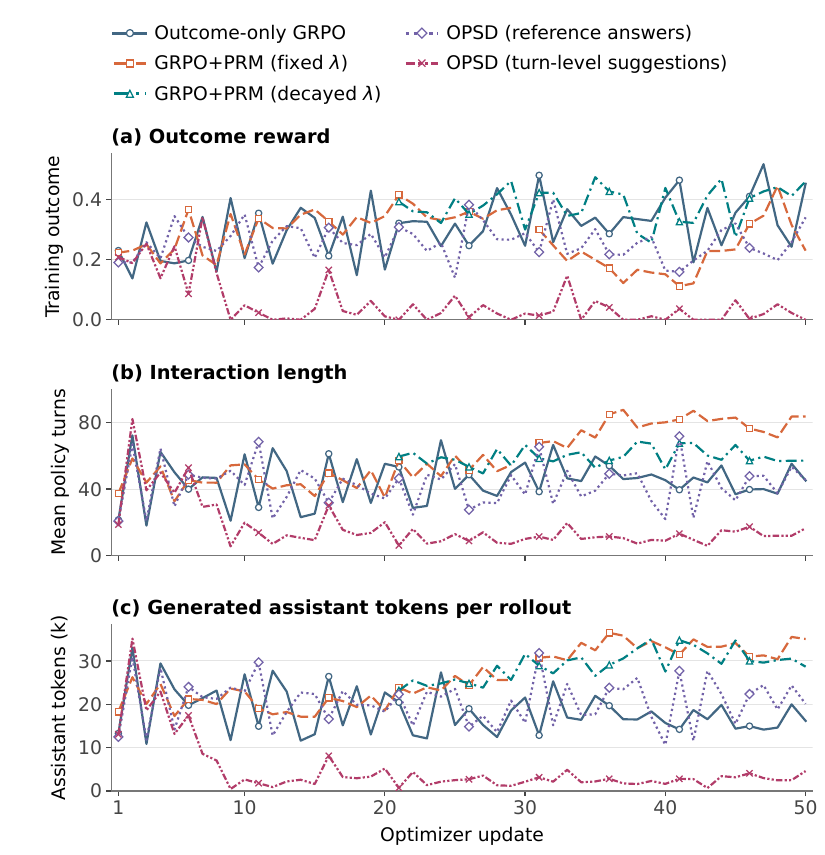}
  \caption{Training dynamics of the five \baseModel{} variants. Each point is an
  unsmoothed, recorded optimizer update: (a) outcome reward on training rollouts,
  (b) policy turns divided by the logged rollout count, and
  (c) logged mean assistant tokens per rollout. Outcome reward is monitored but
  not optimized by \OPSD{}. Records are selected along the final checkpoint
  lineage. The missing fixed-weight update 30 is left as a gap.
  The decayed run starts at update 21 from the fixed-weight update-20 checkpoint;
  its earlier prefix is not replotted. These are individual training runs,
  with different rollout batches, not repeated held-out evaluation scores.}
  \label{fig:appendix-training-trajectories}
\end{figure}

\paragraph{Training dynamics beyond the final evaluation.}
Figure~\ref{fig:appendix-training-trajectories} reports the outcome reward and
interaction length of the training batches for the five \baseModel{} variants.
These are training diagnostics, separate from the repeated held-out evaluation.
Over updates 41--50, \GRPOPRMFixed{} averages 80.5 policy turns per rollout,
compared with 43.8 for \outcomeGRPO{} and 60.7 for \GRPOPRMDecayed{};
the corresponding means of the per-update outcome rewards are 0.257, 0.356,
and 0.394. Thus, the longer trajectories of the fixed-weight run do not coincide
with higher training outcome reward in this interval. These lengths do not
measure truncation or successful task completion.

\begin{figure}[tbp]
  \centering
  \includegraphics[width=\linewidth]{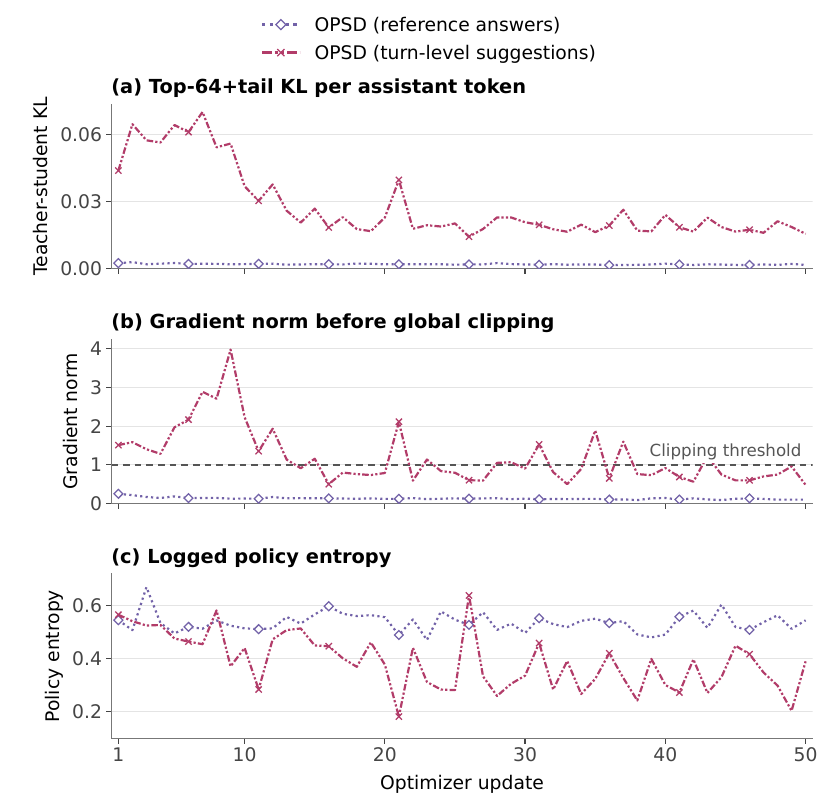}
  \caption{Optimization diagnostics for the two reported \OPSD{} runs over 50
  retained optimizer updates. (a) Token-mean forward KL on the teacher's top 64
  tokens plus a tail class; (b) gradient norm before global clipping, with the
  threshold of 1.0 dashed; and (c) logged policy entropy. Points are unsmoothed
  records from the final checkpoint lineage, with repeated or superseded restart
  attempts excluded. The suggestion-conditioned run has larger KL and gradient
  norms and generally lower later entropy. These diagnostics accompany the
  outcome degradation in Figure~\ref{fig:appendix-training-trajectories}; they
  do not establish a causal mechanism.}
  \label{fig:appendix-training-opsd-stability}
\end{figure}

\paragraph{Observed degradation in \OPSDSuggestions{}.}
The outcome and generation diagnostics make the degradation visible beyond the
training objective. In \OPSDSuggestions{}, mean training outcome reward decreases
from 0.166 over updates 1--10 to 0.020 over updates 41--50, while mean generated
assistant tokens per rollout decrease from 13.9k to 2.9k. For
\OPSDReference{}, the corresponding outcome means are 0.255 and 0.245.
Figure~\ref{fig:appendix-training-opsd-stability} shows larger teacher--student KL
and gradient norms, together with generally lower later policy entropy, for the
suggestion-conditioned run. Its gradient norm exceeds 1.0 on 22 of the 50
retained updates, compared with none for the reference-conditioned run.
These observations document degradation in this configuration; they do not
isolate its cause. The comparisons involve different sampled trajectories,
and the \OPSD{} variants use one rollout per task while the reward-based
variants use eight.

\FloatBarrier

\end{document}